\pdfoutput=1

\documentclass[11pt]{article}

\usepackage{AACL2022/acl} 

\usepackage{times}
\usepackage{latexsym}

\usepackage[T1]{fontenc}

\usepackage[utf8]{inputenc}

\usepackage{microtype}

\usepackage{ownstyles}

\usepackage{booktabs}
\usepackage{nicematrix}
\usepackage{mathtools}
\usepackage{multirow}
\usepackage{makecell}
\usepackage{array}
\usepackage{subcaption}
\usepackage{tabularx,colortbl}
\usepackage{enumitem}
\usepackage{commath}
\usepackage{amsfonts}
\usepackage{soul}
\usepackage{tikz}

\usepackage{float}

\usepackage{multicol} 

\newcommand{\papertitle}{Double Trouble: How to \emph{not} Explain a Text Classifier's Decisions Using Counterfactuals Synthesized by Masked Language Models?}
\title{\papertitle}

\newcommand{\eg}{e.g.\xspace}
\newcommand{\ie}{i.e.\xspace}

\newcommand{\best}[1]{{\textcolor{black}{\bf#1}}}
\newcommand{\worst}[1]{{\textcolor{red}{\bf#1}}}
\definecolor{myGray}{rgb}{0.6,0.6,0.6}
\newcommand{\mask}{\textcolor{myGray}{[MASK]}\xspace}
\newcommand{\unk}{\textcolor{myGray}{[UNK]}\xspace}
\newcommand{\pad}{\textcolor{myGray}{[PAD]}\xspace}
\newcommand{\mybox}[2]{{\color{#1}\fbox{\normalcolor#2}}}
\newcolumntype{M}[1]{>{\centering\arraybackslash}m{#1}}
\renewcommand{\arraystretch}{1.1}

\usepackage{amsmath,amsfonts,bm}

\def\eqref#1{equation~\ref{#1}}

\def\1{\bm{1}}

\def\va{{\bm{a}}}

\def\vx{{\bm{x}}}

\DeclareMathAlphabet{\mathsfit}{\encodingdefault}{\sfdefault}{m}{sl}
\SetMathAlphabet{\mathsfit}{bold}{\encodingdefault}{\sfdefault}{bx}{n}

\def\gH{{\mathcal{H}}}

\def\gV{{\mathcal{V}}}

\def\sR{{\mathbb{R}}}

\newcommand{\E}{\mathbb{E}}

\newcommand{\xti}{\tilde{x}_i}

\newcommand{\bertBase}{BERT$_{\text{\ensuremath{\mathsf{}}}}$\xspace}
\newcommand{\robertaBase}{RoBERTa$_{\text{\ensuremath{\mathsf{}}}}$\xspace}
\newcommand{\ood}{OOD$_{\text{\ensuremath{\mathsf{}}}}$\xspace}
\newcommand{\im}{IM$_{\text{\ensuremath{\mathsf{}}}}$\xspace}
\newcommand{\looEmpty}{LOO$_{\text{\ensuremath{\mathsf{empty}}}}$\xspace}
\newcommand{\looZero}{LOO$_{\text{\ensuremath{\mathsf{zero}}}}$\xspace}
\newcommand{\looUnk}{LOO$_{\text{\ensuremath{\mathsf{unk}}}}$\xspace}
\newcommand{\lime}{LIME$_{\text{\ensuremath{\mathsf{}}}}$\xspace}
\newcommand{\limeBert}{LIME$_{\text{\ensuremath{\mathsf{BERT}}}}$\xspace}
\newcommand{\limeBertSst}{LIME$_{\text{\ensuremath{\mathsf{BERT\_SST2}}}}$\xspace}

\newcommand{\aucRep}{AUC$_{\text{\ensuremath{\mathsf{rep}}}}$\xspace}
\newcommand{\roarBert}{ROAR$_{\text{\ensuremath{\mathsf{BERT}}}}$\xspace}
\newcommand{\roarBertSst}{ROAR$_{\text{\ensuremath{\mathsf{BERT\_SST2}}}}$\xspace}
\newcommand{\deletionBert}{Deletion$_{\text{\ensuremath{\mathsf{BERT}}}}$\xspace}

\newcommand{\subsec}[1]{\noindent{\textbf{#1~~}}}

\newcommand{\class}[1]{{\colorbox{gray!7}{``#1''}}}

\newcommand{\hlborderRed}[1]{\mybox{red}{\strut#1}}
\newcommand{\hlborderGreen}[1]{\mybox{green}{\strut#1}}
\newcommand{\hlborderBlue}[1]{\mybox{blue}{\strut#1}}
\definecolor{myGreen}{rgb}{0.04,0.58,0.22}
\definecolor{myBlue}{rgb}{0.1,0.4,0.56}

\newif\ifcomments

\commentstrue

\ifcomments
\newcommand{\comments}[1]{#1}
\else
\newcommand{\comments}[1]{}
\fi

\title{\papertitle}

\author{Thang M. Pham$^\dagger$ \\
  {\small\texttt{thangpham@auburn.edu}} \\\And
  Trung Bui$^*$ \\
  {\small\texttt{bui@adobe.com}} \\\And
  Long Mai$^*$ \\
  {\small\texttt{mai.t.long88@gmail.com}} \\\AND
  Anh Nguyen$^\dagger$ \\
  \small{\texttt{anh.ng8@gmail.com}} \\\AND
    {\normalfont $^\dagger$Auburn University~~~~~$^*$Adobe Research}
  }

\begin{document}
\maketitle

\begin{abstract}
A principle behind dozens of attribution methods is to take the prediction difference between before-and-after an input feature (here, a token) is removed as its attribution.
A popular Input Marginalization (IM) method \cite{kim-etal-2020-interpretation} uses \bertBase to replace a token, yielding more plausible counterfactuals.
While \citet{kim-etal-2020-interpretation} reported that IM is effective, we find this conclusion not convincing as the \deletionBert metric used in their paper is biased towards IM.
Importantly, this bias exists in Deletion-based metrics, including Insertion, Sufficiency, and Comprehensiveness.
Furthermore, our rigorous evaluation using 6 metrics and 3 datasets finds \textbf{no evidence that IM is better} than a Leave-One-Out (LOO) baseline.
We find two reasons why IM is not better than LOO: (1) deleting a single word from the input only marginally reduces a classifier's accuracy; and (2) a highly predictable word is always given near-zero attribution, regardless of its true importance to the classifier.
In contrast, making Local Interpretable Model-Agnostic Explanations (LIME) counterfactuals more natural via BERT consistently improves LIME accuracy under several RemOve-And-Retrain (ROAR) metrics.

\end{abstract}

\section{Introduction}

\begin{figure}[!ht]
\centering\small 
\setlength\tabcolsep{1pt} 
\begin{tabular}{|l|l|}
\hline
\multicolumn{2}{|l|}{\cellcolor{white} (a) \textbf{SST} -- Groundtruth \& target class: \class{positive}} \\
\hline
S & \makecell[l]{\footnotesize The very definition of the ` small ' movie , but \\ it is a good \colorbox{yellow!100}{\hlborderRed{\strut stepping}} \colorbox{yellow!100}{\hlborderGreen{\strut stone}} \colorbox{yellow!100}{\hlborderBlue{\strut for}} director Sprecher .} \\
& \textbf{0.9793 \hlborderRed{\strut stepping}} ~~~~ \textbf{0.9760 \hlborderGreen{\strut stone}} ~~~~~~~ \textbf{0.8712 \hlborderBlue{\strut for}} \\
& 0.0050 rolling ~~~~~~~~~ 0.0048 stones ~~~~~~~ 0.0860 to \\
& 0.0021 casting ~~~~~~~~  0.0043 point ~~~~~~~~~ 0.0059 , \\

\hline
\hline
IM$_0$ & \makecell[l]{\footnotesize \colorbox{orange!0.383}{\strut The} \colorbox{orange!5.024}{\strut very} \colorbox{orange!4.103}{\strut definition} \colorbox{blue!0.045}{\strut of} \colorbox{orange!21.963}{\strut the} \colorbox{blue!13.406}{\strut `} \colorbox{orange!11.966000000000001}{\strut small} \colorbox{blue!0.128}{\strut '} \colorbox{orange!6.09}{\strut movie} \colorbox{orange!2.252}{\strut ,} \colorbox{orange!83.753}{\strut but} \\  \colorbox{orange!6.625}{\strut it} \colorbox{orange!8.982999999999999}{\strut is} \colorbox{orange!100.0}{\strut a} \colorbox{orange!32.047}{\strut good} 
\colorbox{blue!0.295}{\strut stepping}
\colorbox{orange!2.7969999999999997}{\strut stone}
\colorbox{orange!1.7919999999999998}{\strut for}
\colorbox{orange!40.910000000000004}{\strut director} \colorbox{orange!47.693999999999996}{\strut Sprecher} \colorbox{orange!0.192}{\strut .}} \\

\hline
IM$_{1}$ & \makecell[l]{\footnotesize \colorbox{blue!5.955}{\strut The} \colorbox{blue!100.0}{\strut \textcolor{white}{very}} \colorbox{blue!67.217}{\strut definition} \colorbox{blue!0.482}{\strut of} \colorbox{orange!12.343}{\strut the} \colorbox{blue!13.346}{\strut `} \colorbox{orange!15.136}{\strut small} \colorbox{blue!0.757}{\strut '} \colorbox{blue!22.594}{\strut movie} \colorbox{orange!2.716}{\strut ,} \colorbox{orange!4.729}{\strut but} \\ \colorbox{blue!2.282}{\strut it} \colorbox{blue!54.205999999999996}{\strut is} \colorbox{blue!0.755}{\strut a} \colorbox{blue!30.877}{\strut good} 
\colorbox{blue!0.335}{\strut stepping} 
\colorbox{orange!0.136}{\strut stone} 
\colorbox{orange!1.799}{\strut for}
\colorbox{orange!77.215}{\strut director} \colorbox{orange!43.201}{\strut Sprecher} \colorbox{orange!2.125}{\strut .}} \\

\hline
IM$_{2}$ & \makecell[l]{\footnotesize \colorbox{blue!2.302}{\strut The} \colorbox{blue!39.513}{\strut very} \colorbox{blue!38.714999999999996}{\strut definition} \colorbox{blue!0.455}{\strut of} \colorbox{orange!77.461}{\strut the} \colorbox{blue!53.725}{\strut `} \colorbox{orange!56.117}{\strut small} \colorbox{blue!0.6479999999999999}{\strut '} \colorbox{blue!21.714}{\strut movie} \colorbox{orange!18.365000000000002}{\strut ,} \colorbox{blue!47.382999999999996}{\strut but} \\ \colorbox{orange!6.381}{\strut it} \colorbox{blue!77.398}{\strut is} \colorbox{blue!0.08099999999999999}{\strut a} \colorbox{orange!5.439}{\strut good} 
\colorbox{blue!0.5780000000000001}{\strut stepping}
\colorbox{orange!0.321}{\strut stone}
\colorbox{orange!8.177}{\strut for}
\colorbox{orange!100.0}{\strut director} \colorbox{orange!60.601000000000006}{\strut Sprecher} \colorbox{orange!0.11800000000000001}{\strut .}} \\

\hline
IM$_{3}$ & \makecell[l]{\footnotesize \colorbox{blue!0.296}{\strut The} \colorbox{blue!61.838}{\strut very} \colorbox{blue!45.137}{\strut definition} \colorbox{blue!0.516}{\strut of} \colorbox{blue!30.714000000000002}{\strut the} \colorbox{orange!21.45}{\strut `} \colorbox{orange!22.983999999999998}{\strut small} \colorbox{blue!0.631}{\strut '} \colorbox{blue!31.291999999999998}{\strut movie} \colorbox{orange!0.5}{\strut ,} \colorbox{orange!16.573999999999998}{\strut but} \\ \colorbox{blue!18.883}{\strut it} \colorbox{blue!9.652}{\strut is} \colorbox{blue!3.1879999999999997}{\strut a} \colorbox{blue!100.0}{\strut \textcolor{white}{good}} 
\colorbox{orange!1.315}{\strut stepping}
\colorbox{blue!0.8789999999999999}{\strut stone}
\colorbox{blue!0.718}{\strut for}
\colorbox{orange!71.34700000000001}{\strut director} \colorbox{blue!45.083}{\strut Sprecher} \colorbox{orange!1.702}{\strut .}} \\

\hline
\addlinespace[3mm]
\hline
\multicolumn{2}{|l|}{\cellcolor{white} (b) \textbf{e-SNLI} -- Groundtruth \& target class: \class{contradiction}} \\
\hline
P & \footnotesize A group of people prepare \colorbox{yellow!100}{\hlborderRed{\strut hot}} \colorbox{yellow!100}{\hlborderGreen{\strut air}} \colorbox{yellow!100}{\hlborderBlue{\strut balloons}} for takeoff . \\
& \textbf{0.9997 \hlborderRed{\strut hot}} ~~~~~~~~~~~~~ \textbf{0.9877 \hlborderGreen{\strut air}} ~~~~~~~~~~ \textbf{0.9628 \hlborderBlue{\strut balloons}} \\
& 0.0001 compressed ~ 0.0102 water ~~~~~~~ 0.0282 balloon \\
& 0.0000 open ~~~~~~~~~~~~ 0.0008 helium ~~~~~ 0.0019 engines \\
\hline
H & \footnotesize A group of people prepare \colorbox{yellow!100}{\strut cars} for racing . \\
\hline
\hline

IM$_0$ & \footnotesize \colorbox{orange!0.003}{\strut A} \colorbox{orange!0.022000000000000002}{\strut group} \colorbox{blue!0.001}{\strut of} \colorbox{orange!0.099}{\strut people} \colorbox{orange!0.029}{\strut prepare} 
hot
\colorbox{orange!0.021}{\strut air}
\colorbox{orange!1.328}{\strut balloons}
\colorbox{orange!0.001}{\strut for} \colorbox{orange!1.722}{\strut takeoff} \colorbox{orange!0.002}{\strut .} \\
\hline
& \footnotesize \colorbox{orange!5.118}{\strut A} \colorbox{orange!0.214}{\strut group} \colorbox{orange!-0.0}{\strut of} \colorbox{orange!0.308}{\strut people} \colorbox{blue!0.03}{\strut prepare} \colorbox{orange!100.0}{\strut cars} \colorbox{orange!0.008}{\strut for} \colorbox{orange!21.355}{\strut racing} \colorbox{blue!0.006999999999999999}{\strut .} \\
\hline
\hline

IM$_{1}$ & \footnotesize \colorbox{blue!0.001}{\strut A} \colorbox{blue!0.484}{\strut group} \colorbox{orange!-0.0}{\strut of} \colorbox{blue!0.623}{\strut people} \colorbox{orange!0.922}{\strut prepare} 
hot
\colorbox{orange!0.097}{\strut air}
\colorbox{orange!1.1320000000000001}{\strut balloons}
\colorbox{orange!0.09}{\strut for} \colorbox{blue!12.665000000000001}{\strut takeoff} \colorbox{orange!0.214}{\strut .} \\
\hline
& \footnotesize \colorbox{blue!12.450999999999999}{\strut A} \colorbox{blue!2.156}{\strut group} \colorbox{blue!0.003}{\strut of} \colorbox{blue!0.189}{\strut people} \colorbox{orange!2.521}{\strut prepare} \colorbox{orange!80.414}{\strut cars} \colorbox{blue!0.038}{\strut for} \colorbox{blue!100.0}{\strut \textcolor{white}{racing}} \colorbox{orange!0.251}{\strut .} \\
\hline
\hline

IM$_{2}$ & \footnotesize \colorbox{blue!0.002}{\strut A} \colorbox{blue!0.5640000000000001}{\strut group} \colorbox{blue!0.001}{\strut of} \colorbox{blue!0.122}{\strut people} \colorbox{orange!0.152}{\strut prepare} 
\colorbox{blue!0.002}{\strut hot}
\colorbox{orange!0.215}{\strut air}
\colorbox{orange!0.9400000000000001}{\strut balloons}
\colorbox{orange!0.1}{\strut for} \colorbox{orange!9.59}{\strut takeoff} \colorbox{blue!0.638}{\strut .} \\
\hline
& \footnotesize \colorbox{blue!5.737}{\strut A} \colorbox{blue!2.5839999999999996}{\strut group} \colorbox{blue!0.003}{\strut of} \colorbox{blue!0.22799999999999998}{\strut people} \colorbox{blue!1.1769999999999998}{\strut prepare} \colorbox{blue!45.73}{\strut cars} \colorbox{blue!0.066}{\strut for} \colorbox{blue!100.0}{\strut \textcolor{white}{racing}} \colorbox{blue!0.065}{\strut .} \\
\hline
\hline

IM$_{3}$ & \footnotesize \colorbox{blue!0.003}{\strut A} \colorbox{blue!0.054}{\strut group} \colorbox{orange!-0.0}{\strut of} \colorbox{orange!0.167}{\strut people} \colorbox{orange!0.08099999999999999}{\strut prepare} 
\colorbox{blue!0.001}{\strut hot}
\colorbox{orange!0.08499999999999999}{\strut air}
\colorbox{orange!0.565}{\strut balloons}
\colorbox{orange!0.024}{\strut for} \colorbox{blue!6.0760000000000005}{\strut takeoff} \colorbox{blue!0.013}{\strut .} \\
\hline
& \footnotesize \colorbox{blue!1.17}{\strut A} \colorbox{blue!0.42700000000000005}{\strut group} \colorbox{blue!0.001}{\strut of} \colorbox{blue!0.013999999999999999}{\strut people} \colorbox{blue!0.04}{\strut prepare} \colorbox{blue!5.0889999999999995}{\strut cars} \colorbox{blue!0.022000000000000002}{\strut for} \colorbox{blue!100.0}{\strut \textcolor{white}{racing}} \colorbox{blue!0.013999999999999999}{\strut .} \\
\hline
\end{tabular}
\caption{
\textbf{
By design, IM erroneously assigns near-zero attribution to highly-predictable words.}
Color map:
\colorbox{blue!100}{\strut \textcolor{white}{negative -1}}, neutral 0, \colorbox{orange!100}{\strut positive +1}.
Many words labeled \colorbox{yellow!100}{\strut important} by humans such as 
``\hlborderRed{\strut stepping}'', 
``\hlborderGreen{\strut stone}'' (a) or 
``\hlborderRed{\strut hot}'',
``\hlborderGreen{\strut air}''
(b) are always given near-zero attribution by IM (because they are highly predictable by BERT, e.g. 0.9793 for \hlborderRed{\strut stepping}) regardless of the classifier.
Even when randomizing the classifier's weights three times, the IM attribution of these words remains unchanged at near zero (IM$_1$ to IM$_3$).
Therefore, when marginalizing over the top-$k$ BERT candidates (\eg, ``stepping'', ``rolling'', ``casting''), the IM attribution for low-entropy words tends to zero, leading to heatmaps that are biased, less accurate, and less plausible than \looEmpty.
}
\label{fig:top1_duplicate}
\end{figure}

Feature attribution maps (AMs), \ie highlights indicating the importance of each input token w.r.t. a classifier's decision, can help improve \emph{human accuracy} on downstream tasks including detecting fake movie reviews \cite{lai2019human} or identifying biases in text classifiers \cite{liu2019incorporating}.
Many Leave-One-Out (LOO) methods compute the attribution of an input token by measuring the prediction changes after substituting that token's embedding with zeros \cite{li2016understanding,jin2020towards} or \unk \cite{kim-etal-2020-interpretation}.
That is, deleting or replacing features is the underlying principle of at least 25 attribution methods \cite{covert2020feature}.

Based on the evidence in computer vision \cite{bansal2020sam,zhang2019should}, prior works in NLP \emph{hypothesized}
that removing a word from an input text forms out-of-distribution (OOD) inputs that yield erroneous AMs \cite{kim-etal-2020-interpretation,harbecke-alt-2020-considering} or AMs inconsistent with human's perception of causality \cite{hase2021out}.
To generate plausible counterfactuals, two teams of researchers \cite{kim-etal-2020-interpretation,harbecke-alt-2020-considering} proposed Input Marginalization (IM), i.e. replace a word using \bertBase \cite{devlin2019bert} and compute an average prediction difference by marginalizing over all predicted words.
\citet{kim-etal-2020-interpretation} claimed that IM yields more accurate AMs than the baselines that replace words by \unk or zeros but their quantitative results were reported for only \emph{one}\footnote{No \emph{quantitative} results on SNLI, only SST-2.} dataset and \emph{one} evaluation metric.




In this paper, we re-assess their claim by, first, reproducing their IM results\footnote{Code and pre-trained models are available at 
\url{https://github.com/anguyen8/im}.
}, 
and then rigorously evaluate whether improving the realism of counterfactuals improves two attribution methods (LOO and LIME).
On a diverse set of \emph{three} datasets and \emph{six} metrics, we find that:


\begin{enumerate}
    \item The \deletionBert metric in \citet{kim-etal-2020-interpretation} is biased towards IM as both use BERT to replace words (Sec.~\ref{sec:contradiction}).
    In contrast, the vanilla Deletion metric \cite{arras2017explaining} favors the \looEmpty baseline as both delete words.
    This bias causes a \textbf{false conclusion} that IM is better than LOO baselines in \citet{kim-etal-2020-interpretation} and also \textbf{exists in other Deletion variants}, \eg, Insertion \cite{arras2017explaining}, Sufficiency, and Comprehensiveness \cite{deyoung-etal-2020-eraser}.

    \item We find \textbf{no evidence that IM is better} than a simple \looEmpty on any of the following four state-of-the-art AM evaluation metrics (which exclude the biased Deletion \& \deletionBert): ROAR, \roarBert \cite{hooker2019benchmark} (Sec.~\ref{sec:roar}), comparison against human annotations (Sec.~\ref{sec:human_annotations}), and sanity check \cite{adebayo2018sanity} (Sec.~\ref{sec:sanity_check}).

    \item We argue that IM is not effective in practice because: (1) deleting a single word from an input has only a marginal effect on classification accuracy (Sec.~\ref{sec:ood}); and (2) given a \emph{perfect}, masked language model $G$, IM would still be \textbf{unfaithful} because highly predictable words according to $G$, \eg ``hot'', ``air'' in Fig.\ref{fig:top1_duplicate}, are always assigned near-zero attribution in IM \emph{regardless} of how important they are to the classifier (Sec.~\ref{sec:bert_top1}).

    \item To further test the main idea of IM, we integrate BERT into LIME \cite{ribeiro2016should} to \emph{replace} multiple words (instead of deleting) in an input sequence, making LIME counterfactuals more realistic.
    We find this technique to improve LIME consistently under multiple ROAR-based metrics, but not under comparison against human annotations (Sec.~\ref{sec:lime_vs_lime_bert}).

\end{enumerate}

To our knowledge, our work is the first to thoroughly study the effectiveness of IM in NLP in both settings of replacing a single word (LOO) and multiple words (LIME).
Importantly, we find improvement in the latter but not the former setting.


\section{Methods and Related Work}
\label{sec:methods}


Let $f:\sR^{n \times d} \to [0,1]$ be a text classifier that maps a sequence $\vx$ of $n$ token embeddings, each of size $d$, onto a confidence score of an output label.
An attribution function $A$ takes three inputs---a sequence $\vx$, the model $f$, and a set of hyperparameters $\gH$---and outputs a vector $\va = A (f, \vx, \gH) \in [-1,1]^{n}$.
Here, the explanation $\va$ associates each input token $x_i$ to a scalar $a_i \in [-1,1]$, indicating how much $x_i$ contributes for or against the target label.

\paragraph{Leave-One-Out} (LOO) is a well-known method \cite{li2016understanding,robnik2008explaining,jin2020towards} for estimating the attribution $a_i$ by computing the prediction-difference after a token $x_i$ is left out of the input $\vx$, creating a shorter sequence $\vx_{-i}$:

\begin{align}
    a_i = f(\vx) - f(\vx_{-i}) \label{eq:pred_diff}
\end{align}

 
Under \citet{pearl2009causality} causal framework, the attribution $a_i$ in Eq.~\ref{eq:pred_diff} relies on a single, unrealistic counterfactual $\vx_{-i}$ and thus is a biased estimate of the individual treatment effect (ITE):

\begin{align}
    \text{ITE} &= f(\vx) - \E[ f(\vx) \mid do (T = 0)] \label{eq:ITE}
\end{align}

\noindent where the binary treatment $T$, here, is to keep or ``realistically remove'' the token $x_i$ (\ie $T = 1$ or $0$) in the input $\vx$, prior to the computation of $f(\vx)$.


\paragraph{Perturbation techniques}
In computer vision (CV), earlier attribution methods erase a feature by replacing it with (a) zeros \cite{zeiler2014visualizing,ribeiro2016should}; (b) random noise \cite{dabkowski2017real,lundberg2017unified}; or (c) blurred versions of the original content \cite{fong2019understanding}.
Yet, these perturbation methods produce unrealistic counterfactuals that make AMs more unstable and less accurate \cite{bansal2020sam}.

Recent works proposed to simulate the $do(T = 0)$ operator using an image inpainter.
However, they either generated unnatural counterfactuals \cite{chang2018explaining,goyal2019explaining} or only a single, plausible counterfactual per example \cite{agarwal2020explaining}.

\paragraph{Input marginalization (IM)} 
In NLP, \im offers the closest estimate of the ITE.
IM computes the $\E[.]$ term in Eq.~\ref{eq:ITE} by marginalizing over many plausible counterfactuals generated by \bertBase: 

\begin{align}
    \E[f(\vx) \mid &~do(T=0)] \nonumber\\
    &=\sum_{\xti\in{\gV}} p(\xti| \vx_{-i}) \cdot f(\vx_{-i}, \xti) \label{eq:expectation_term}
\end{align}

\noindent where $\xti$ is a token suggested by \bertBase (e.g., ``\hlborderRed{\strut hot}'', ``compressed'', or ``open'' in Fig.~\ref{fig:top1_duplicate}) with a likelihood of $p(\xti|\vx_{-i})$ to replace the masked token $x_i$.
$\gV$ is the \bertBase vocabulary of 30,522 tokens.
$f(\vx_{-i}, \xti)$ is the classification probability when token $x_i$ in the original input is replaced with $\xti$.

IM attribution is in the $\log$ space:

\begin{align}
    a_\text{IM} &= \text{log-odds} (f(\vx)) \nonumber\\ 
    &~~~~~- \text{log-odds} ( \E [f(\vx) \mid do(T = 0)] ) \label{eq:marginalization}
\end{align}

\noindent where $\text{log-odds} (p) = \log_2 (p / (1 - p))$.

As computing the expectation in Eq.~\ref{eq:expectation_term} over BERT's $\sim$30K-word vocabulary is prohibitively slow, IM authors only marginalized over the words that have a likelihood $\geq 10^{-5}$.
We are \emph{able to reproduce} the \im results of \citet{kim-etal-2020-interpretation} by taking only the top-10 words.
That is, using the top-10 words or all words of likelihood $\geq 10{-5}$ yields slightly different numbers but the same conclusions (Sec.~\ref{sec:appendix_original_vs_modified_im}).
Thus, we marginalize over the top-10 for all experiments.
Note that under BERT, the top-10 tokens, on average, already account for $81\%$, $90\%$, and $92\%$ of the probability mass for SST-2, e-SNLI, \& MultiRC, respectively.

\paragraph{BERT}
Like \citet{kim-etal-2020-interpretation}, we use a pre-trained \bertBase ``base'', uncased model \citep{devlin2019bert}, from \citet{huggingface2020pretrained},  to fill in a \mask token to generate counterfactuals in IM.

\paragraph{LIME}
Based on the idea of IM, we also integrate \bertBase into \lime, which originally masks out multiple tokens at once to compute attribution.
\lime generates a set of randomly masked versions of the input, and the attribution of a token $x_i$, is effectively the mean classification probability over all the masked inputs when $x_i$ is not masked out.
On average, each vanilla \lime counterfactual has 50\% of tokens taken out, yielding text often with large syntactic and grammatical errors.

\paragraph{\limeBert} 
We use BERT to replace multiple masked tokens\footnote{We find replacing all tokens at once or one at a time to produce similar \limeBert results.} in each masked sentence generated by LIME to construct more plausible counterfactuals.
However, for each word, we only use the top-1 highest-likelihood token given by BERT instead of marginalizing over multiple tokens because (1) the full marginalization is prohibitively slow; and (2) the top-1 token already carries most of the weight ($p \geq 0.81$; see Table~\ref{table:bert_top1}).



\section{Experiment framework}

\subsection{Three datasets}


We select a diverse set of three classification datasets that enable us to (1) compare with the results reported by \citet{kim-etal-2020-interpretation}; and (2) assess AMs on six evaluation metrics (described in Sec.~\ref{sec:metrics}).
These three tasks span from sentiment analysis (SST-2), natural language inference (e-SNLI) to question answering (MultiRC), covering a wide range of sequence length ($\sim$20, 24, and 299 tokens per example, respectively).
SST-2 and e-SNLI were the two datasets where \citet{kim-etal-2020-interpretation} found \im to be superior to LOO baselines.





\paragraph{SST} Stanford Sentiment Treebank \citep{socher2013recursive} is a dataset of $\sim$12K RottenTomato movie-review \emph{sentences}, which contain human-annotated sentiment annotations for phrases.
Each phrase and sentence in SST is assigned a sentiment score $\in  [0,1]$ (0 = negative, 0.5 = neutral, 1 = positive).

\paragraph{SST-2} has $\sim$70K SST examples (including both phrases and sentences) where the regression scores per example were binarized to form a binary classification task \cite{socher2013recursive}.

\paragraph{e-SNLI}
A 3-way classification task of detecting whether the relation between a premise and a hypothesis is entailment, neutral or contradiction \cite{snli:emnlp2015}.
e-SNLI has 569K instances of (input, label, explanation) where the explanations are crowd-sourced \cite{NEURIPS2018_4c7a167b}.


\paragraph{MultiRC}
Multi-sentence Reading Comprehension \citep{khashabi-etal-2018-looking} is a multiple-choice question-answering task that provides multiple input sentences as well as a question and asks the model to select one or multiple correct answer sentences.
MultiRC has $\sim$6K examples with human-annotated highlights at the sentence level.

\subsection{Classifiers}

Following \citet{kim-etal-2020-interpretation,harbecke-alt-2020-considering,hase2021out}, we test \im and LOO baselines in explaining BERT-based classifiers.

For each task, we train a classifier by fine-tuning the entire model, which consists of a classification layer on top of the pre-trained BERT (described in Sec.~\ref{sec:methods}).
The dev-set top-1 accuracy scores of our SST-2, e-SNLI, \& MultiRC classifiers are 92.66\%, 90.92\%, and 69.10\%, respectively.
On the SST binarized dev-set, which contains only sentences, the SST-2-trained classifier's accuracy is 87.83\%.


\paragraph{Hyperparameters} 
Following the training scheme of HuggingFace, we fine-tune all classifiers for 3 epochs using Adam optimizer \citep{kingma2014adam} with a learning rate of 0.00002, $\beta_1$ = 0.9, $\beta_2$ = 0.999, $\epsilon = 10^{-8}$.
A batch size of 32 and a max sequence length of 128 are used for SST-2 and e-SNLI while these hyperparameters for MultiRC are 8 and 512, respectively. 
Dropout with a probability of 0.1 is applied to all layers.
Each model was trained on an NVIDIA 1080Ti GPU.

\subsection{Six evaluation metrics}
\label{sec:metrics}



As there are \emph{no groundtruth} explanations in XAI, we use six common metrics to rigorously assess IM’s effectiveness.
For each classifier, we evaluate the AMs generated for all dev-set examples.

\subsec{Deletion} is similar to ``Comprehensiveness'' \cite{deyoung-etal-2020-eraser} and is based on the idea that deleting a token of higher importance from the input should cause a larger drop in the output confidence score.
We take the original input and delete one token at a time until 20\% of the tokens in the input is deleted.
A more accurate explanation is expected to have a lower Area Under the output-probability Curve (AUC) \cite{arras2017explaining}.

\subsec{\deletionBert} a.k.a. \aucRep in \citet{kim-etal-2020-interpretation}, is a Deletion variant where a given token is replaced by a BERT top-1 suggestion instead of an empty string.
\deletionBert was proposed to minimize the OOD-ness of samples (introduced by deleting words in the vanilla Deletion metric), \ie akin to integrating BERT into LOO to create IM.

\subsec{RemOve And Retrain (ROAR)}
To avoid a potential \ood generalization issue caused by the Deletion metric, a common alternative is to retrain the classifier on these modified inputs (where $N$\% of the highest-attribution words are deleted) and measure its accuracy drop \cite{hooker2019benchmark}.
A more faithful attribution method is supposed to lead to a re-trained classifier of lower accuracy as the more important words have been deleted from training examples.
For completeness, we also implement \textbf{\roarBert}, which uses BERT to replace the highest-attribution tokens\footnote{The chance that a sentence remains unchanged after BERT replacement is low, $\leq$ 1\%.} instead of deleting them without replacement in ROAR.




\subsec{Agreement with human-annotated highlights}
In both CV and NLP, a common AM evaluation metric is to assess the agreement between AMs and human annotations \cite{wiegreffe2021teach}.
The idea is that as text classifiers well predict the human labels of an input text, their explanations, \ie AMs, should also highlight the tokens that humans deem indicative of the groundtruth label.

Because human annotators only label the tokens supportive of a label (\eg Fig.~\ref{fig:loo_vs_im_sst}), when comparing AMs with human annotations, we zero out the \colorbox{blue!100}{\strut \textcolor{white}{negative}} values in AMs.
Following \citet{zhou2016learning}, we binarize a resulting AM at an optimal threshold $\tau$ in order to compare it with human-annotated highlights under Precision@1.

\subsec{Sanity check} \cite{adebayo2018sanity} is a well-known metric for testing insensitivity (\ie bias) of attribution methods w.r.t. model parameters.
For ease of interpretation, we compute the \% change of per-word attribution values in \emph{sign} and \emph{magnitude} as we randomize the classification layer's weights.
A better attribution method is expected to be more sensitive to the classifier's weight randomization.

\section{Bias of Deletion metric and its variants}
\label{sec:contradiction}


In explaining SST-2 classifiers, we successfully reproduce the \aucRep results reported in \citet{kim-etal-2020-interpretation}, \ie \im outperformed \looZero and \looUnk, which were implemented by replacing a word with the \pad and \unk token of BERT, respectively (Table~\ref{table:auc_vs_auc_rep}).
However, we hypothesize that \deletionBert is biased towards IM as both use BERT to replace words, yielding a false sense of IM effectiveness reported in \citet{kim-etal-2020-interpretation}.

To test this hypothesis, we add another baseline of \looEmpty, which was \emph{not} included in \citet{kim-etal-2020-interpretation}, \ie erasing a token from the input without replacement (Eq.~\ref{eq:pred_diff}), mirroring the original Deletion metric.
To compare with \im, all LOO methods in this paper are also in the log-odds space.



\subsec{Results} 
Interestingly, we find that, under Deletion, on both SST-2 and e-SNLI, \im \emph{underperformed all} three LOO baselines and that \looEmpty is the highest-performing method (Table~\ref{table:auc_vs_auc_rep}a).
In contrast, \im is the best method under \deletionBert.

Re-running the same experiment but sampling replacement words from RoBERTa (instead of BERT), we find the same finding that \looEmpty is the best under Deletion while IM is the best under \deletionBert (Table~\ref{table:auc_vs_auc_rep}b).






\begin{table}[ht]
\centering\small
\setlength\tabcolsep{2.3pt}
\begin{tabular}{llcccc}
\toprule
Task & Metrics $\downarrow$ & \im & \looZero & \looUnk & \looEmpty \\
\toprule
\rowcolor{gray!25} \multicolumn{6}{c}{(a) \bertBase} \\
\cmidrule{0-5}
\multirow{2}{*}{SST-2} & Deletion & \worst{0.4732} & 0.4374 & 0.4464 & \best{0.4241} \\
\cmidrule(l{2pt}r{2pt}){2-6}
& \deletionBert & \best{0.4922} & 0.4970 & 0.5047 & \worst{0.5065} \\
\cmidrule{0-5}
\multirow{2}{*}{e-SNLI} & Deletion & \worst{0.3912} & 0.2798 & 0.3742 & \best{0.2506} \\
\cmidrule(l{2pt}r{2pt}){2-6}
& \deletionBert & \best{0.2816} & 0.3240 & \worst{0.3636} & 0.3328 \\
\cmidrule{0-5}
\multicolumn{6}{c}{\cellcolor{gray!25} (b) \robertaBase} \\
\cmidrule{0-5}
\multirow{2}{*}{SST-2} & Deletion & \worst{0.4981} & 0.4524 & 0.4595 & \best{0.4416} \\
\cmidrule(l{2pt}r{2pt}){2-6}
& \deletionBert & \best{0.4798} & 0.5037 & \worst{0.5087} & {0.4998} \\
\bottomrule
\end{tabular}
\caption{
IM is the \best{best} method under \deletionBert, as reported in \citet{kim-etal-2020-interpretation}, but the \worst{worst} under Deletion.
Both metrics measure AUC (lower is better).
}
\label{table:auc_vs_auc_rep}
\end{table}

To our knowledge, our work is \textbf{the first to document this bias} of the Deletion metric \textbf{widely used in the literature} \cite{hase2021out,wiegreffe2021teach,arras2017explaining}.
This bias, in principle, also \textbf{exists in other Deletion variants} including Insertion \cite{arras2017explaining}, Sufficiency, and Comprehensiveness \cite{deyoung-etal-2020-eraser}.

\section{No evidence that IM is better than LOO}
\label{sec:unbiased_metrics}

To avoid the critical bias of Deletion and \deletionBert, we further compare IM and LOO on \textbf{four} common metrics that are not Deletion-based.

\subsection{Under ROAR and \roarBert, IM is on-par with or worse than \looEmpty}
\label{sec:roar}

A lower AUC under Deletion may be the artifact of the classifier misbehaving under the distribution shift when one or multiple input words are deleted.
ROAR \cite{hooker2019benchmark} was designed to ameliorate this issue by re-training the classifier on a modified training-set (where the top $N\%$ highest-attribution tokens in each example are deleted) before evaluating their accuracy.

To more objectively assess \im, we use ROAR and \roarBert metrics to compare \im vs. \looEmpty (\ie the best LOO variant in Table~\ref{table:auc_vs_auc_rep}).
\subsec{Experiment}
For both \im and \looEmpty, we generate AMs for every example in the SST-2 train and dev sets, and remove $N\%$ highest-attribution tokens per example to create new train and dev sets.
We train 5 models on the new training set and evaluate them on the new dev set.
We repeat ROAR and \roarBert with $N \in \{10, 20, 30\}$.\footnote{We do not use $N \geq 40$ because:
(1) according to SST human annotations, only 37\% of the tokens per example are labeled \class{important} (Table \ref{table:loo_im_human_coverage}c); and
(2) SST-2 examples are short and may contain as few as 4 tokens per example.}


\subsec{Results}
As more tokens are removed (\ie $N$ increases), the mean accuracy of 5 models gradually decreases (Table~\ref{table:roar}; from 92.66\% to $\sim$67\%).
Under both ROAR and \roarBert, the models trained on the new training set derived from \looEmpty AMs often obtain lower (\ie better) mean accuracy than those of \im (Table~\ref{table:roar}a vs. b).
At $N = 10\%$ under ROAR, \textbf{\looEmpty outperforms IM} (Table~\ref{table:roar}; 74.59 vs. 76.22), which is statistically significant (2-sample $t$-test, $p = 0.037$).
In all other cases, the difference between IM vs. \looEmpty is not statistically significant.

In sum, under both ROAR and \roarBert, IM is \emph{not more faithful} than \looEmpty.

\begin{table*}[ht] 
\centering\small
\setlength\tabcolsep{3.5pt}
\begin{NiceTabular}{lcccc|ccc}
\toprule
\multicolumn{2}{l}{Accuracy in \% (lower is better)} & \multicolumn{3}{c}{ROAR} & \multicolumn{3}{c}{\roarBert} \\
\cmidrule(l{2pt}r{2pt}){3-8}
Method & $N = 0$\% & 10\% & 20\% & 30\% & 10\% & 20\% & 30\% \\
\toprule
(a) \looEmpty & 92.62 $\pm$ 0.30 & \textbf{74.59} $\pm$ 0.78 & \textbf{68.94} $\pm$ 1.46 & {67.89} $\pm$ 0.79 & \textbf{76.79} $\pm$ 0.56 & 71.95 $\pm$ 0.75 & \textbf{67.62} $\pm$ 1.16 \\
\cmidrule(l{2pt}r{2pt}){0-7}
(b) \im & 92.62 $\pm$ 0.30 & 76.22 $\pm$ 1.18 & 70.07 $\pm$ 0.69 & \textbf{66.54} $\pm$ 1.89 & 77.36 $\pm$ 0.90 & \textbf{71.56} $\pm$ 1.55 & 67.68 $\pm$ 0.96 \\
\cmidrule(l{2pt}r{2pt}){0-7} 
(c) Random & 92.62 $\pm$ 0.30 & 89.22 $\pm$ 0.53 & 87.75 $\pm$ 0.19 & 85.62 $\pm$ 0.53 & 89.38 $\pm$ 0.47 & 88.23 $\pm$ 0.31 & 85.21 $\pm$ 0.47 \\
\hline
\cmidrule{0-7}
(d) $t$-test p-value & N/A & \textbf{0.0370} & 0.1740 & 0.1974 & 0.2672 & 0.6312 & 0.9245 \\
\bottomrule
\end{NiceTabular}
\caption{
Dev-set mean accuracy (\%) of 5 models trained on the new SST-2 examples where $N$\% of highest-attribution words per example are removed (\ie ROAR) or replaced via BERT (\ie \roarBert).
On average, under both metrics, \looEmpty (a) is slightly better, \ie lower mean accuracy, than \im (b). 
Notably, \looEmpty statistically significantly outperforms \im under ROAR at $N=10$\% (2-sample $t$-test; $p = 0.037$) (d).
Both \looEmpty and \im substantially outperform a random baseline (c) that considers $N\%$ random tokens important.
}
\label{table:roar} 
\end{table*}

\begin{table*}[ht]
\centering\small
\setlength\tabcolsep{2.2pt}
\begin{NiceTabular}{ccc|ccc|cc|cc|cc}
\toprule
Metric $\uparrow$ & \multicolumn{5}{c}{(a) SST} & \multicolumn{2}{c}{(b) e-SNLI {~~L2}} & \multicolumn{2}{c}{(c) e-SNLI {~~L3}} & \multicolumn{2}{c}{(d) MultiRC} \\
\cmidrule(l{2pt}r{2pt}){2-12}
{\tiny Higher is better} & \im & \looEmpty & \lime & \limeBert & \limeBertSst & \im & \looEmpty & \im & \looEmpty & \im & \looEmpty \\ 

\toprule
IoU & 0.2377 & \textbf{0.2756} & 0.3193 & 0.3170 & 0.3127 & 0.3316 & \textbf{0.3415} & 0.2811 & \textbf{0.3411} & 0.0437 & \textbf{0.0887} \\
\cmidrule(l{2pt}r{2pt}){1-12}
precision & \textbf{0.5129} & 0.4760 & 0.4831 & 0.4629 & 0.4671 & 0.4599 & \textbf{0.4867} & 0.3814 & \textbf{0.4687} & 0.1784 & \textbf{0.1940} \\
\cmidrule(l{2pt}r{2pt}){1-12}
recall & 0.5245 & \textbf{0.6077} & 0.6882 & 0.7000 & 0.6886 & 0.6085 & \textbf{0.6158} & 0.5699 & \textbf{0.5875} & 0.0630 & \textbf{0.2876} \\
\cmidrule(l{2pt}r{2pt}){1-12}
F1 & 0.5186 & \textbf{0.5338} & 0.5677 & 0.5573 & 0.5566 & 0.5239 & \textbf{0.5437} & 0.4570 & \textbf{0.5214} & 0.0931 & \textbf{0.2317} \\
\bottomrule
\end{NiceTabular}
\caption{
Compared to \im, \looEmpty is substantially more consistent with human annotations over all three datasets.
Note that the gap between \looEmpty and \im is $\sim$$3\times$ wider when comparing AMs with the e-SNLI tokens that at least three annotators label \class{important} (\ie L3), compared to L2 (higher is better).
\limeBert explanations are slightly less consistent with human highlights than those of LIME (a) despite their counterfactuals are more realistic.
}
\label{table:human_annotations}
\end{table*}

\subsection{\looEmpty aligns significantly better with human annotations than IM}
\label{sec:human_annotations}

Following \citet{wiegreffe2021teach}, to increase our understanding of the differences between \looEmpty and IM, we compare the two methods against the human-annotated highlights for SST, e-SNLI, and MultiRC.




\subsec{Annotation preprocessing}
To control for quality, we preprocess the human annotations in each dataset as the following.
In SST, where each sentence has multiple phrases labeled with a sentiment score $\in [0, 1]$ (0.5 being the \class{neutral} midpoint), we only use the phrases that have high-confidence sentiment scores, \ie $\leq 0.3$ (for \class{negative}) or $\geq 0.7$ (for \class{positive}).
Also, we do not use the annotated phrases that are too long, \ie, longer than 50\% of the sentence length.


Each token in an e-SNLI example are labeled \class{important} by between 0--3 annotators.
To filter out noise, we only use the tokens that are highlighted by \emph{at least} two or three annotators (hereafter ``L2'' and ``L3'' subsets, respectively).



A MultiRC example contains a question and a paragraph where each sentence is labeled \class{important} or \class{unimportant} to the groundtruth answer (Fig.~\ref{fig:qualitative_examples_multirc_1a}).
We convert these sentence-level highlights into token-level highlights to compare them with the binarized AMs of IM and \looEmpty.

\subsec{Experiment}
We run \im and \looEmpty on the BERT-based classifiers on the dev set of SST, e-SNLI, and MultiRC.
All AMs generated are binarized using a threshold $\tau \in \{ 0.05x \mid 0 < x < 20 \text{ and } x \in \mathbb{N} \}$.
We compute the average IoU, precision, recall, and F1 over pairs of (human binary map, binarized AM) and report the results at the optimal $\tau$ of each explanation method.
For both \looEmpty and \im, $\tau = 0.1$ on SNLI-L2 and 0.05 on both SST-2 and MultiRC.
On SNLI-L3, $\tau$ is 0.40 and 0.45 for \looEmpty and \im, respectively.





\subsec{SST results}
We found that \looEmpty aligns better with human highlights than \im (Figs.~\ref{fig:loo_vs_im_sst} \&~\ref{fig:qualitative_examples_sst_2c}).
\looEmpty outperforms \im in both F1 and IoU scores (Table~\ref{table:human_annotations}a; 0.2756 vs 0.2377) with a notably large recall gap (0.6077 vs. 0.5245).

\begin{figure}[ht]
\centering\small 
\setlength\tabcolsep{5pt}

\begin{tabular}{|c|l|}
\hline
\multicolumn{2}{|l|}{\cellcolor{white} \textbf{SST}~ Groundtruth \& Prediction: \class{positive} movie review} \\
\hline

Input &
\makecell[l]{\footnotesize Mr. Tsai \colorbox{yellow!100}{\strut is} \colorbox{yellow!100}{\strut a} \colorbox{yellow!100}{\strut very} \colorbox{yellow!100}{\strut original} \colorbox{yellow!100}{\strut artist} \colorbox{yellow!100}{\strut in} \colorbox{yellow!100}{\strut his} \colorbox{yellow!100}{\strut medium} , \\ and What Time Is It There ?} \\
\hline
\hline
IM & \makecell[l]{\footnotesize Mr. \colorbox{ProcessBlue!100}{\strut Tsai} is a very \colorbox{ProcessBlue!100}{\strut original} artist in his \colorbox{ProcessBlue!100}{\strut medium} \colorbox{ProcessBlue!100}{\strut ,} \\ \colorbox{ProcessBlue!100}{\strut and} What Time Is It \colorbox{ProcessBlue!100}{\strut There} ?} \\
\hline
& IoU: 0.17, precision: 0.33, recall: 0.25 \\
\hline
\hline
LOO & \makecell[l]{\footnotesize Mr. \colorbox{WildStrawberry!100}{\strut Tsai} \colorbox{WildStrawberry!100}{\strut is} \colorbox{WildStrawberry!100}{\strut a} \colorbox{WildStrawberry!100}{\strut very} \colorbox{WildStrawberry!100}{\strut original} \colorbox{WildStrawberry!100}{\strut artist} \colorbox{WildStrawberry!100}{\strut in} \colorbox{WildStrawberry!100}{\strut his} \colorbox{WildStrawberry!100}{\strut medium} , \\ and What Time \colorbox{WildStrawberry!100}{\strut Is} It There ?} \\
\hline
& IoU: \textbf{0.80}, precision: \textbf{0.80}, recall: \textbf{1.00} \\
\hline
\end{tabular}

\caption{
\looEmpty binarized \colorbox{WildStrawberry!100}{\strut attribution maps} align better with \colorbox{yellow!100}{\strut human} highlights than \colorbox{ProcessBlue!100}{\strut IM maps}.
}
\label{fig:loo_vs_im_sst}
\end{figure}





\subsec{e-SNLI and MultiRC results}
Similarly, in both tasks, \looEmpty explanations are more consistent with human highlights than \im explanations under all four metrics (see Table \ref{table:human_annotations}b--d and qualitative examples in Figs.~\ref{fig:loo_vs_im_esnli} \&~\ref{fig:qualitative_examples_esnli_2b}--\ref{fig:loo_vs_im_multirc2}).

Remarkably, in MultiRC where each example is substantially longer ($\sim$299 tokens per example) than those in the other tasks, the recall and F1 scores of \looEmpty is, respectively, 2$\times$ and 4$\times$ higher than those of \im (see Table \ref{table:human_annotations}).


\begin{figure}[ht]
\centering\small 
\setlength\tabcolsep{2pt} 
\begin{tabular}{|c|l|}
\hline
\multicolumn{2}{|l|}{\cellcolor{white} \textbf{e-SNLI} example.~ Groundtruth \& Prediction: \class{entailment}} \\
\hline
P & \makecell[l]{\footnotesize Two men dressed in black \colorbox{yellow!100}{\strut practicing} \colorbox{yellow!100}{\strut martial} \colorbox{yellow!100}{\strut arts} \\ on a gym floor .} \\
\hline
H & \footnotesize Two men are \colorbox{yellow!100}{\strut doing} \colorbox{yellow!100}{\strut martial} \colorbox{yellow!100}{\strut arts} . \\
\hline
\hline
IM & \makecell[l]{\footnotesize \colorbox{ProcessBlue!100}{\strut Two} \colorbox{ProcessBlue!100}{\strut men} dressed in black practicing martial arts \\ on a gym floor .} \\
& \footnotesize \colorbox{ProcessBlue!100}{\strut Two} \colorbox{ProcessBlue!100}{\strut men} \colorbox{ProcessBlue!100}{\strut are} \colorbox{ProcessBlue!100}{\strut doing} martial arts . \\
\hline
& IoU: 0.09, precision: 0.17, recall: 0.16 \\
\hline
\hline
LOO & \makecell[l]{\footnotesize \colorbox{WildStrawberry!100}{\strut Two} \colorbox{WildStrawberry!100}{\strut men} dressed in black \colorbox{WildStrawberry!100}{\strut practicing} \colorbox{WildStrawberry!100}{\strut martial} \colorbox{WildStrawberry!100}{\strut arts} \\ on a gym floor .} \\
& \footnotesize Two \colorbox{WildStrawberry!100}{\strut men} are \colorbox{WildStrawberry!100}{\strut doing} \colorbox{WildStrawberry!100}{\strut martial} arts \colorbox{WildStrawberry!100}{\strut .} \\
\hline
& IoU: \textbf{0.50}, precision: \textbf{0.56}, recall: \textbf{0.83} \\
\hline
\end{tabular}
\caption{
\looEmpty important \colorbox{WildStrawberry!100}{\strut words} are in a stronger agreement with \colorbox{yellow!100}{\strut human} highlights than \colorbox{ProcessBlue!100}{\strut IM important words}.
Each e-SNLI example contains a pair of premise (P) and hypothesis (H).
}
\label{fig:loo_vs_im_esnli}
\end{figure}

\subsection{IM is insensitive to model randomization}
\label{sec:sanity_check}

\citet{adebayo2018sanity} found that many attribution methods can be surprisingly biased, \ie \emph{insensitive} to even randomization of the classifier's parameters.
Here, we test the degree of insensitivity of IM when the last classification layer of BERT-based classifiers is randomly re-initialized.
We use three SST-2 classifiers and three e-SNLI classifiers.



Surprisingly, IM is consistently worse than \looEmpty, \ie more insensitive to classifier randomization.
That is, on average, the IM attribution of a word changes signs (from positive to negative or vice versa) less frequently, e.g. 62.27\% of the time, compared to 71.41\% for \looEmpty on SST-2 (Table~\ref{table:sanity}a).
The average change in attribution \emph{magnitude} of IM is also $\sim$1.5$\times$ smaller than that of \looEmpty (Table~\ref{table:sanity}b).

For example, the IM attribution scores of \hlborderRed{\strut hot}, \hlborderGreen{\strut air} or \hlborderBlue{\strut balloons} in Fig.~\ref{fig:top1_duplicate}  remain consistently \textbf{unchanged near-zero even when the classifier is randomized three times}.
That is, each of these three words is $\sim$100\% predictable by BERT given the other two words (Fig.~\ref{fig:top1_duplicate}b; IM$_1$ to IM$_3$) and, hence, will be assigned a near-zero attribute by IM (by construction, via Eqn.~\ref{eq:expectation_term} \& \ref{eq:marginalization}) regardless of how important these words actually are to the classifier.
Statistically, this is a major issue because across SST, e-SNLI, and MultiRC, we find BERT to correctly predict the missing word $\sim$49, 60, 65\% of the time, respectively (Sec.~\ref{sec:low_magnitude_coverage}).
And that the average likelihood score of a top-1 exact-match token is high, $\sim$0.81--0.86 (Sec.~\ref{sec:bert_top1}), causing the highly predicted words (\eg, \hlborderRed{\strut hot}) to always be assigned low attribution regardless of their true importance to the classifier.

We find this insensitivity to be a major, \textbf{theoretical flaw of IM} in explaining a classifier's decision at the \emph{word} level.
By analyzing the overlap between IM explanations and human highlights (generated in experiments in Sec.~\ref{sec:human_annotations}), we find consistent results that IM explanations have \textbf{significantly smaller attribution magnitude} per token (Sec.~\ref{sec:low_magnitude_coverage}) and \textbf{substantially lower recall than LOO} (Sec.~\ref{sec:bert_top1}). 

\subsection{Classification accuracy only drops marginally when one token is deleted}
\label{sec:ood}

Our previous results show that replacing \emph{a single word} by BERT (instead of deleting) in IM creates more realistic inputs but actually hurts the AM quality w.r.t. LOO.
This result interestingly contradicts the prior conclusions \cite{kim-etal-2020-interpretation,harbecke-alt-2020-considering} and assumptions \cite{hase2021out} of the superiority of IM over LOO.


To understand why using more plausible counterfactuals did not improve AM explainability, we assess the $\Delta$ drop in classification accuracy when a word is deleted (\ie, \looEmpty samples; Fig.~\ref{fig:ood}) and the $\Delta$ when a word is replaced via BERT (\ie IM samples).

\subsec{Results}
Across SST, e-SNLI, and MultiRC, the accuracy scores of classifiers only drop marginally $\sim$1--4 points (Table \ref{table:ood}) when a single token is deleted.
See Figs.~\ref{fig:ood}~\&~\ref{fig:qualitative_examples_esnli_3a} for qualitative examples showing that deleting a single token hardly changes the predicted label.
Whether a word is removed or replaced by BERT is almost unimportant in tasks with long examples such as MultiRC (Table \ref{table:ood}; 1.10 and 0.24).
In sum, we do not find the unnaturalness of LOO samples to substantially hurt model performance, questioning the need raised in \cite{hase2021out,harbecke-alt-2020-considering,kim-etal-2020-interpretation} for realistic counterfactuals.

\begin{table}[ht] 
\centering\small
\setlength\tabcolsep{3pt}
\begin{tabular}{lccc}
\toprule
$\Delta$ drop in accuracy (\%) & \multicolumn{1}{c}{SST} & \multicolumn{1}{c}{e-SNLI} & \multicolumn{1}{c}{MultiRC} \\
\toprule
(a) LOO~~~~(1-token deleted) & 3.52 & 4.92 & 1.10 \\
\cmidrule(l{2pt}r{2pt}){1-4}
(b) \im \hfill(1-token replaced) & 2.20 & 4.86 & 0.24 \\
\cmidrule(l{2pt}r{2pt}){1-4}
(c) \lime \hfill(many tokens deleted) & 16.38 & 25.74 & 17.85 \\
\bottomrule
\end{tabular}
\caption{
The dev-set accuracies on SST, e-SNLI and MultiRC (87.83\%, 90.92\%, and 69.10\%, respectively) only drop marginally when a single token is deleted (a) or replaced using BERT (b).
In contrast, LIME samples cause the classification accuracy to drop substantially (\eg 16.38 points on SST).
}
\label{table:ood}
\end{table}

\section{Replacing (instead of deleting) \emph{multiple} words can improve explanations}
\label{sec:lime_vs_lime_bert}

We find that deleting a single word only marginally affects classification accuracy.
Yet, deleting $\sim$50\% of words, \ie following LIME's counterfactual sampling scheme, actually substantially reduces classification accuracy, \eg $-$16.38 point on SST and $-$25.74 point on e-SNLI (Table \ref{table:ood}c).
Therefore, it is interesting to test whether the core idea of harnessing BERT to replace words has merits in improving LIME whose counterfactuals are extremely OOD due to many missing words.

\subsection{\limeBert attribution maps are \emph{not} more aligned with human annotations}
\label{sec:lime_annotations}

Similar to Sec.~\ref{sec:human_annotations}, here, we compare \lime and \limeBert AMs with human SST annotations (avoiding the Deletion-derived metrics due to their bias described in Sec.~\ref{sec:contradiction}).

\paragraph{Experiment}
We use the default hyperparameters of the original LIME \cite{lime2021implement} for both \lime and \limeBert.
The number of counterfactual samples was 1,000 per example.


\subsec{Results}
Although \limeBert counterfactuals are more natural, the derived AMs are surprisingly less plausible to human than those generated by the original \lime.
That is, compared to human annotations in SST, \limeBert's IoU, precision and F1 scores are all slightly worse than those of \lime  (Table~\ref{table:human_annotations}a).
Consistent with the IM vs. \looEmpty comparison in Sec.~\ref{sec:human_annotations}, {replacing one or more words (instead of deleting them) using BERT in LIME generates AMs that are similarly or less aligned with humans}.

To minimize the possibility that the pre-trained BERT is suboptimal in predicting missing words on SST-2, we also finetune BERT using the mask-language modeling objective on SST-2 (see details in Sec.~\ref{sec:train_bert_sst2}) and repeat the experiment in this section.
Yet, interestingly, we find the above conclusion to not change (Table~\ref{table:human_annotations}a; \limeBertSst is worse than LIME).
In sum, for both LOO and LIME, we find \textbf{no evidence that using realistic counterfactuals from BERT causes AMs to be more consistent with words that are labeled \class{important} by humans}.

\subsection{\limeBert consistently outperforms LIME under three ROAR metrics}
\label{sec:LIME}

To thoroughly test the idea of using BERT-based counterfactuals in improving LIME explanations, we follow Sec.~\ref{sec:roar} and compare \limeBert and LIME under three ROAR metrics: (1) ROAR; (2) \roarBert; and (3) \roarBertSst, \ie which uses the BERT finetuned on SST-2 to generate training data.

\subsec{Experiment} Similar to the previous section, we take the dev set of SST-2 and generate a LIME AM and a LIME-BERT AM for each SST-2 example.
For \roarBertSst, we re-use the BERT finetuned on SST-2 described in Sec.~\ref{sec:lime_annotations}.

\begin{figure}[ht]
\centering
\includegraphics[width=\linewidth]{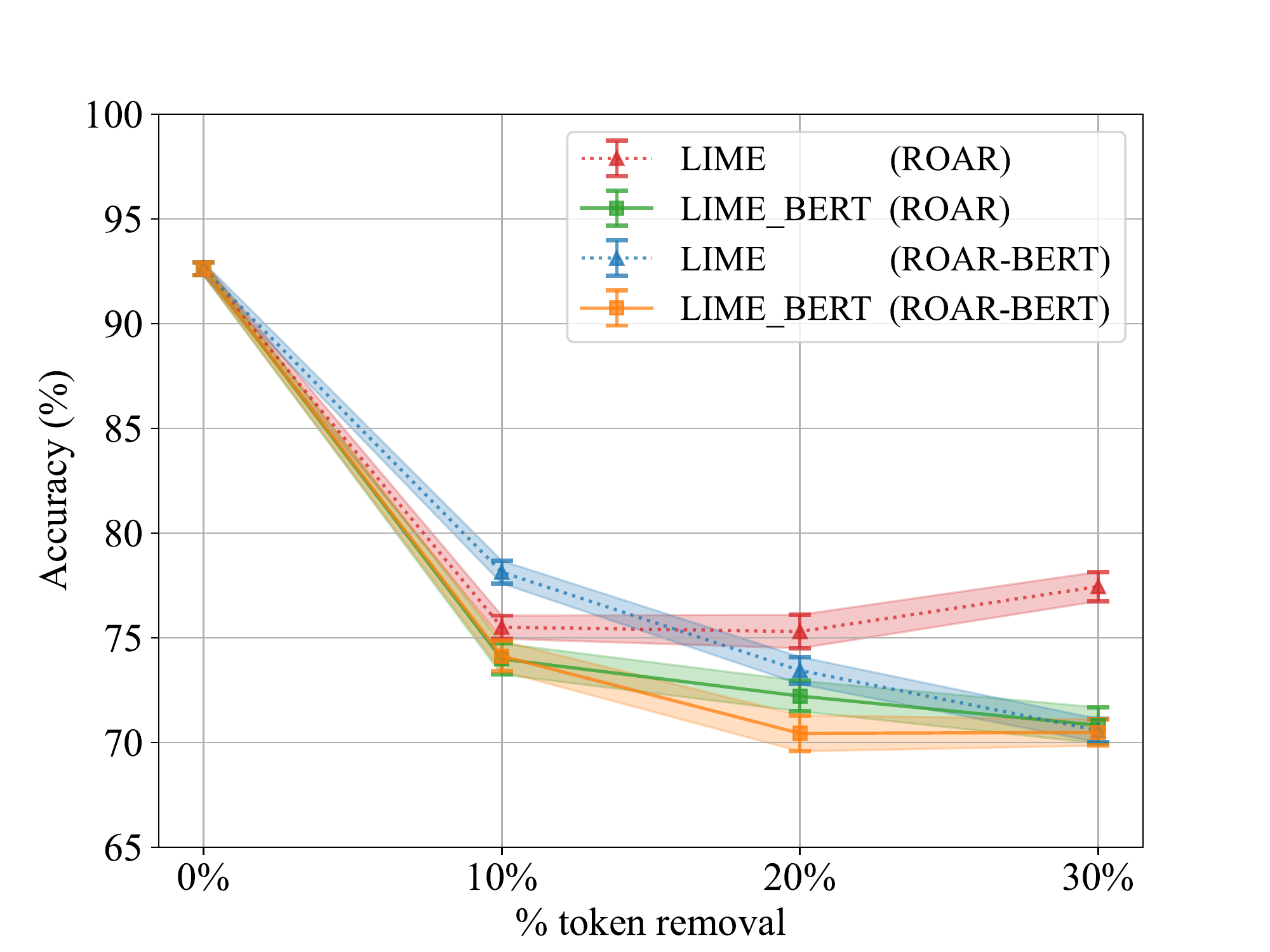}
\caption{
\limeBert slightly, but consistently outperforms LIME when evaluated under either ROAR or \roarBert.
The each point in the $y$-axis shows the mean accuracy of five different classifiers.
See more results supporting the same conclusion in Table~\ref{table:lime_vs_limebert}.
}
\label{fig:lime_vs_limebert}
\end{figure}

\subsec{Results}
Interestingly, we find that \limeBert slightly, but consistently outperforms LIME via all three ROAR metrics tested (Fig.~\ref{fig:lime_vs_limebert}; dotted lines are above solid lines).
That is, \limeBert tend to highlight more discriminative tokens in the text than LIME, yielding a better ROAR performance (\ie lower accuracy in Table~\ref{table:lime_vs_limebert}).
This result is consistent across all three settings of removing 10\%, 20\%, and 30\% most important words, and when using either pre-trained BERT or BERT finetuned on SST-2.




\section{Discussion and Conclusion}
\label{sec:discussion}



We find in Sec.~\ref{sec:sanity_check} that IM is highly insensitive to classifier's changes because, by design, it always assigns near-zero attribution to highly-predictable words $x_i$ regardless of their true importance to a target classifier.
A solution may be to leave such $x_i$ token out of the marginalization (Eq.~\ref{eq:expectation_term}), \ie only marginalizing over the other tokens suggested by BERT.
However, these other replacement tokens altogether have a sum likelihood of 0.
That is, replacing token $x_i$ by zero-probability tokens (\ie truly implausible) would effectively generate OOD text, which, in turn is not desired \cite{hase2021out}.

Our results in Sec.~\ref{sec:LIME} suggests that IM might be more useful at the \emph{phrase} level \cite{jin2020towards} instead of \emph{word} level as deleting a set of contiguous words has a larger effect to the classifier predictions.

In sum, for the first time, we find that the popular idea of harnessing BERT to generate realistic counterfactuals \cite{hase2021out,harbecke-alt-2020-considering,kim-etal-2020-interpretation} does not actually improve upon a simple \looEmpty in practice as an \looEmpty counterfactual only has a single word deleted.
In contrast, we observe more expected benefits of this technique in improving methods like LIME that has counterfactuals that are extremely syntactically erroneous when multiple words are often deleted.


\subsection*{Acknowledgments}
We thank Michael Alcorn, Qi Li, and Peijie Chen for helpful feedback on the early results.
We also thank the anonymous reviewers for their detailed and constructive criticisms that helped us improve the manuscript.
AN is supported by the National Science Foundation (grant 1850117; 2145767) and donations from Adobe Research and the NaphCare Charitable Foundation.


\clearpage
\bibliography{references}
\bibliographystyle{AACL2022/acl_natbib}

\appendix


\newcommand{\toptitlebar}{
    \hrule height 4pt
    \vskip 0.25in
    \vskip -\parskip%
}
\newcommand{\bottomtitlebar}{
    \vskip 0.29in
    \vskip -\parskip%
    \hrule height 1pt
    \vskip 0.09in%
}

\newcommand{\beginsupplementary}{%
    \setcounter{table}{0}
    \renewcommand{\thetable}{A\arabic{table}}%
    \setcounter{figure}{0}
    \renewcommand{\thefigure}{A\arabic{figure}}%
    \setcounter{section}{0}
}

\newcommand{\suptitle}{Appendix for:\\\papertitle}

\beginsupplementary
 



\clearpage
{\Large \bf Appendix\par}

\section{\im explanations have smaller attribution magnitude per token and lower word coverage}
\label{sec:low_magnitude_coverage}

To further understand the impact of the fact that BERT tends to not change a to-remove token (Sec.~\ref{sec:bert_top1}), here, we quantify the magnitude of attribution given by \im and its coverage of important words in an example.

\paragraph{Smaller attribution magnitude} 
Across three datasets, the average absolute values of attribution scores (which are $\in [-1, 1])$ of \im are not higher than that of \looEmpty (Table \ref{table:magnitude}).
Especially in MultiRC, \im average attribution magnitude is 4.5$\times$ lower than that of \looEmpty (0.02 vs 0.09).

\begin{table}[ht] 
\centering\small
\setlength\tabcolsep{6.5pt}
\begin{tabular}{lccc}
\toprule
Method & \multicolumn{1}{c}{SST} & \multicolumn{1}{c}{e-SNLI} & \multicolumn{1}{c}{MultiRC} \\
\toprule
\looEmpty & \textbf{0.22} $\pm$ 0.27 & 0.15 $\pm$ 0.24 & \textbf{0.09} $\pm$ 0.09 \\
\cmidrule(l{2pt}r{2pt}){1-4}
\im & 0.17 $\pm$ 0.27 & 0.15 $\pm$ 0.27 & 0.02 $\pm$ 0.09 \\
\bottomrule
\end{tabular}
\caption{
The average absolute value of attribution scores per token of \looEmpty is consistently higher than that of \im.
}
\label{table:magnitude}
\end{table}

\paragraph{Lower word coverage}
We define \emph{coverage} as the average number of highlighted tokens per example (\eg Fig.~\ref{fig:top1_duplicate}) after binarizing a heatmap at the method's optimal threshold.

The coverage of \looEmpty is much higher than that of \im on SST (40\% vs 30\%) and MultiRC examples (27\% vs 6\%), which is consistent with the higher \emph{recall} of \looEmpty (Table \ref{table:loo_im_human_coverage}; a vs. b).
For e-SNLI, although \im has higher coverage than \looEmpty (14\% vs. 10\%), the coverage of \looEmpty is closer to the human coverage (9\%).
That is, \im assigns high attribution incorrectly to many words, resulting in a substantially lower \emph{precision} than \looEmpty, according to e-SNLI L3 annotations (Table~\ref{table:human_annotations}b; 0.3814 vs. 0.4687).

\begin{table}[ht] 
\centering\small
\setlength\tabcolsep{6pt}
\begin{tabular}{lrrrr}
\toprule
Explanations & \multicolumn{1}{c}{SST} & \multicolumn{2}{c}{e-SNLI} & \multicolumn{1}{c}{MultiRC} \\
\cmidrule(l{2pt}r{2pt}){3-4}
generated by & \multicolumn{1}{c}{} & \multicolumn{1}{c}{L2} & \multicolumn{1}{c}{L3} & \multicolumn{1}{c}{} \\
\toprule
(a) \looEmpty & 40\% & 19\% & 10\% & 27\% \\
\cmidrule(l{2pt}r{2pt}){1-5}
(b) \im & 30\% & 21\% & 14\% & 6\% \\
\cmidrule(l{2pt}r{2pt}){1-5}
(c) Human & 37\% & 18\% & 9\% & 16\% \\
\hline
\cmidrule{1-5}
\# tokens per example & 20 & \multicolumn{2}{c}{24} & 299 \\
\bottomrule
\end{tabular}
\caption{
Compared to \im, the coverage of \looEmpty is closer to the coverage of human explanations.
}
\label{table:loo_im_human_coverage}
\end{table}

In sum, \textbf{chaining our results together}, we found BERT to often replace a token $x_i$ by an exact-match with a high likelihood (Sec.~\ref{sec:bert_top1}), which sets a low empirical upper-bound on attribution values of \im, causing \im explanations to have smaller attribution magnitude.
As the result, after binarization, fewer tokens remain highlighted in \im binary maps (\eg Fig.~\ref{fig:loo_vs_im_esnli}).

\section{By design, IM always assigns near-zero attribution to high-likelihood words regardless of classifiers}
\label{sec:bert_top1}

We observe that \im scores a substantially lower recall compared to \looEmpty (\eg 0.0630 vs. 0.2876; Table~\ref{table:human_annotations}d).
That is, \im tends to incorrectly assign too small of attribution to important tokens.
Here, we test whether this low-recall issue is because BERT is highly accurate at predicting a single missing word from the remaining text and therefore assigns a high likelihood to such words in Eq.~\ref{eq:expectation_term}, causing low IM attribution in Eq.~\ref{eq:ITE}.

\paragraph{Experiment}
For each example in all three datasets, we replaced a single word by BERT's top-1 highest-likelihood token and measured its likelihood and whether the replacement is the same as the original word.

\paragraph{Results}
Across SST, e-SNLI, and MultiRC, the top-1 BERT token matches exactly the original word $\sim$49, 60, 65\% of the time, respectively (Table \ref{table:bert_top1}a).
This increasing trend of exact-match frequency (from SST, e-SNLI $\to$ MultiRC) is consistent with the example length in these three datasets, which is understandable as a word tends to be more predictable given a longer context. 
Among the tokens that human annotators label \class{important}, this exact-match frequency is similarly high (Table \ref{table:bert_top1}b).
Importantly, the average likelihood score of a top-1 exact-match token is high, $\sim$0.81--0.86 (Table \ref{table:bert_top1}c).
See Fig.~\ref{fig:top1_duplicate} \& Figs.~\ref{fig:qualitative_examples_sst_1b}--\ref{fig:qualitative_examples_multirc_1b} for qualitative examples.

\begin{table}[ht] 
\centering\small
\setlength\tabcolsep{4.5pt}
\begin{tabular}{llccc}
\toprule

\% exact-match (uncased) & \multicolumn{1}{c}{SST} & \multicolumn{1}{c}{e-SNLI} & \multicolumn{1}{c}{MultiRC} \\
\toprule
(a) over all tokens  & 48.94 & 59.43 & 64.78 \\
\cmidrule(l{2pt}r{2pt}){1-4}
\multirow{2}{*}{\makecell[c]{(b) over human highlights}}  & \multirow{2}{*}{41.25} & \multirow{2}{*}{42.74} & \multirow{2}{*}{68.55} \\
&  &  & \\
\hline
\cmidrule{1-4}
(c) Top-1 word's likelihood & 0.8229 & 0.8146 & 0.8556 \\
\bottomrule
\end{tabular}
\caption{
Top-1 likelihood scores (c) are the mean likelihood given by BERT for the top-1 predicted words that exactly match the original words (a).
}
\label{table:bert_top1}
\end{table}

Our findings are aligned with \im's low recall.
That is, if BERT fills in an exact-match $\xti$ for an original word $x_i$, the prediction difference for this replacement $\xti$ will be 0 in Eq.~\ref{eq:marginalization}.
Furthermore, a high likelihood of $\sim$0.81 for $\xti$ sets an \textbf{empirical upper-bound of 0.19 for the attribution of the word} $x_i$, which explains the insensitivity of IM to classifier randomization (Fig.~\ref{fig:top1_duplicate}; IM$_1$ to IM$_3$).

The analysis here is also consistent with our additional findings that IM attribution tends to be smaller than that of \looEmpty and therefore leads to heatmaps of lower coverage of the words labeled \class{important} by humans (see Sec.~\ref{sec:low_magnitude_coverage}).

\section{Train BERT as masked language model on SST-2 to help filling in missing words}
\label{sec:train_bert_sst2}

Integrating pre-trained BERT into \lime helps improve LIME explanations under two ROAR metrics (Sec.~\ref{sec:lime_vs_lime_bert}).
However, the pre-trained BERT might be suboptimal for the cloze task on SST-2 sentences as it was pre-trained on Wikipedia and BookCorpus.
Therefore, here, we take the pre-trained BERT, and finetune it on SST-2 training set using the masked language modeling objective.
That is, we aim to test whether having a more specialized BERT would improve LIME results even further.

\paragraph{Training details} We follow the hyperparameters by \cite{huggingface2020pretrained} and use Adam optimizer \cite{kingma2014adam} with a learning rate of 0.00005, $\beta_1$ = 0.9, $\beta_2$ = 0.999, $\epsilon = 10^{-8}$, a batch size of 8, max sequence length of 512 and the ratio of tokens to mask of 0.15.
We finetune the pre-trained BERT on SST-2 \cite{socher-etal-2013-recursive} train set and select the best model using the dev set.

\paragraph{Results}
On the SST-2 test set of 1,821 examples that contain 35,025 tokens in total, the cross-entropy loss of pre-trained BERT and BERT-SST2 are 3.50 $\pm$ 4.58 and 3.29 $\pm$ 4.40, respectively.
That is, our BERT finetuned on SST-2 is better than pre-trained BERT at predicting missing words in SST-2 sentences.


\section{Comparison between original and modified version of Input Marginalization}
\label{sec:appendix_original_vs_modified_im}

We follow \citet{kim-etal-2020-interpretation} to reproduce results of the original Input Marginalization (IM) (Table~\ref{table:appendix_im_im_modified}a--b). 
To reduce the time complexity of Input Marginalization, we propose a modified version (IM-top10) by only marginalizing over the top-10 tokens sampled from BERT rather than using all tokens of likelihood $\geq$ a threshold $\sigma = 10^{-5}$.
We find that IM-top10 has comparable performance to that of the original IM (0.4732 vs. 0.4783; Table~\ref{table:appendix_im_im_modified}c).
Our IM-top10 quantitative results are also close to the original numbers reported in \citet{kim-etal-2020-interpretation} (0.4922 vs. 0.4972; Table~\ref{table:appendix_im_im_modified}).

\begin{table}[ht] 
\centering\small
\setlength\tabcolsep{3pt}
\begin{tabular}{lccc}
\toprule
Metrics $\downarrow$ & a. \im (\emph{reported} in & b. \im  & c. \im-top10 \\
& \citet{kim-etal-2020-interpretation}) & \multicolumn{2}{c}{(Our reproduction)}  \\
\toprule
Deletion & n/a & 0.4783 & {0.4732} \\
\cmidrule(l{2pt}r{2pt}){1-4}
\deletionBert & 0.4972 & {0.4824} & 0.4922 \\
\bottomrule
\end{tabular}
\caption{
The approximation in  of IM-top10 compared to the original IM under two metrics on SST-2 task.
Both metrics measure AUC (lower is better).
}
\label{table:appendix_im_im_modified}
\end{table}

We also find high qualitative similarity between heatmaps produced by two versions: IM vs. IM-top10 (Figs.~\ref{fig:qualitative_examples_sst2_1a}--5). 
The average Pearson correlation score across the SST-2 8720-example test set is fairly high ($\rho = 0.7224$).
Thus, we use IM-top10 for all experiments in this paper.

\begin{figure*}[ht]
\centering
\def\arraystretch{1.5}
\resizebox{\textwidth}{!}{%
\begin{tabular}{|c|c|c|c|c|c|c|c|c|c|c|}
\hline
\multicolumn{11}{|l|}{\cellcolor{white} \textbf{SST-2} example.~ Groundtruth: \class{positive} \& ~Prediction: \class{positive} (Confidence: 0.9996)} \\
\hline
\multirow{2}{*}{\makecell[c]{IM}} & \colorbox{orange!36.076}{\strut among} & \colorbox{orange!0.23500000000000001}{\strut the} & \colorbox{orange!10.764999999999999}{\strut year} & \colorbox{orange!4.4510000000000005}{\strut 's} & \colorbox{orange!20.564}{\strut most} & \colorbox{orange!100.0}{\strut intriguing} & \colorbox{orange!38.582}{\strut explorations} & \colorbox{orange!30.567}{\strut of} & \colorbox{blue!6.235}{\strut alientation} & \colorbox{blue!0.052}{\strut .} \\
 & 1.815 & 0.0118 & 0.54158 & 0.22394 & 1.03458 & 5.03105 & 1.94109 & 1.53783 & -0.31367 & -0.0026 \\ \hline

\multirow{2}{*}{\makecell[c]{IM \\ \textit{modified}}} & \colorbox{orange!46.055}{\strut among} & \colorbox{orange!0.622}{\strut the} & \colorbox{orange!6.022}{\strut year} & \colorbox{orange!9.017999999999999}{\strut 's} & \colorbox{orange!28.087}{\strut most} & \colorbox{orange!100.0}{\strut intriguing} & \colorbox{orange!72.538}{\strut explorations} & \colorbox{orange!40.068}{\strut of} & \colorbox{blue!6.114}{\strut alientation} & \colorbox{orange!0.249}{\strut .} \\
 & 2.64685 & 0.03574 & 0.34608 & 0.51827 & 1.61421 & 5.74711 & 4.16886 & 2.30276 & -0.35139 & 0.01431 \\ \hline

\end{tabular}%
}
\caption{
Color map:
\colorbox{blue!100}{\strut \textcolor{white}{negative -1}}, neutral 0, \colorbox{orange!100}{\strut positive +1}.
Attribution maps derived from both versions of IM have a high Pearson correlation $\rho = 0.988$.
}
\label{fig:qualitative_examples_sst2_1a}
\end{figure*}

\begin{figure*}[ht]
\centering
\def\arraystretch{1.5}
\resizebox{\textwidth}{!}{%
\begin{tabular}{|c|c|c|c|c|c|c|c|c|c|}
\hline
\multicolumn{10}{|l|}{\cellcolor{white} \textbf{SST-2} example.~ Groundtruth: \class{positive} \& ~Prediction: \class{positive} (Confidence: 0.9994)} \\
\hline
\multirow{2}{*}{\makecell[c]{IM}} & \colorbox{orange!17.468}{\strut a} & \colorbox{orange!100.0}{\strut solid} & \colorbox{orange!47.351}{\strut examination} & \colorbox{blue!0.244}{\strut of} & \colorbox{orange!2.325}{\strut the} & \colorbox{blue!6.514}{\strut male} & \colorbox{orange!2.6839999999999997}{\strut midlife} & \colorbox{orange!21.07}{\strut crisis} & \colorbox{orange!19.900000000000002}{\strut .} \\
 & 1.07654 & 6.16288 & 2.91817 & -0.01502 & 0.14328 & -0.40143 & 0.1654 & 1.29851 & 1.2264 \\ \hline

\multirow{2}{*}{\makecell[c]{IM \\ \textit{modified}}} & \colorbox{orange!31.365}{\strut a} & \colorbox{orange!100.0}{\strut solid} & \colorbox{orange!49.537}{\strut examination} & \colorbox{orange!0.013999999999999999}{\strut of} & \colorbox{orange!0.346}{\strut the} & \colorbox{blue!1.9640000000000002}{\strut male} & \colorbox{orange!1.149}{\strut midlife} & \colorbox{orange!18.993}{\strut crisis} & \colorbox{orange!1.016}{\strut .} \\
 & 1.83532 & 5.85144 & 2.89864 & 0.00083 & 0.02024 & -0.11491 & 0.06725 & 1.11138 & 0.05947 \\ \hline

\end{tabular}%
}
\caption{
Color map:
\colorbox{blue!100}{\strut \textcolor{white}{negative -1}}, neutral 0, \colorbox{orange!100}{\strut positive +1}.
Attribution maps derived from both versions of IM have a high Pearson correlation $\rho = 0.917$.
}
\label{fig:qualitative_examples_sst2_1b}
\end{figure*}

\begin{figure*}[ht]
\centering
\def\arraystretch{1.5}
\resizebox{\textwidth}{!}{%
\begin{tabular}{|c|c|c|c|c|c|c|c|c|c|}
\hline
\multicolumn{10}{|l|}{\cellcolor{white} \textbf{SST-2} example.~ Groundtruth: \class{negative} \& ~Prediction: \class{positive} (Confidence: 0.9868)} \\
\hline
\multirow{2}{*}{\makecell[c]{IM}} & \colorbox{orange!79.07}{\strut rarely} & \colorbox{orange!11.77}{\strut has} & \colorbox{blue!25.738}{\strut leukemia} & \colorbox{blue!1.9980000000000002}{\strut looked} & \colorbox{orange!7.099}{\strut so} & \colorbox{orange!100.0}{\strut shimmering} & \colorbox{orange!41.808}{\strut and} & \colorbox{orange!1.882}{\strut benign} & \colorbox{orange!0.61}{\strut .} \\
 & 6.62645 & 0.98643 & -2.15698 & -0.16744 & 0.59491 & 8.38053 & 3.50372 & 0.15773 & 0.05112 \\ \hline

\multirow{2}{*}{\makecell[c]{IM \\ \textit{modified}}} & \colorbox{orange!35.636}{\strut rarely} & \colorbox{orange!6.715999999999999}{\strut has} & \colorbox{blue!37.785000000000004}{\strut leukemia} & \colorbox{blue!2.3890000000000002}{\strut looked} & \colorbox{orange!3.4410000000000003}{\strut so} & \colorbox{orange!100.0}{\strut shimmering} & \colorbox{orange!43.718}{\strut and} & \colorbox{orange!3.005}{\strut benign} & \colorbox{orange!0.563}{\strut .} \\
 & 3.11005 & 0.58616 & -3.29759 & -0.20848 & 0.3003 & 8.72728 & 3.81542 & 0.26226 & 0.04914 \\ \hline

\end{tabular}%
}
\caption{
Color map:
\colorbox{blue!100}{\strut \textcolor{white}{negative -1}}, neutral 0, \colorbox{orange!100}{\strut positive +1}.
Attribution maps derived from both versions of IM have a high Pearson correlation $\rho = 0.983$.
}
\label{fig:qualitative_examples_sst2_1c}
\end{figure*}

\begin{figure*}[ht]
\centering
\def\arraystretch{1.5}
\resizebox{\textwidth}{!}{%
\begin{tabular}{|c|c|c|c|c|c|c|c|c|c|c|c|c|c|c|}
\hline
\multicolumn{15}{|l|}{\cellcolor{white} \textbf{SST-2} example.~ Groundtruth: \class{negative} \& ~Prediction: \class{negative} (Confidence: 0.9950)} \\
\hline
\multirow{2}{*}{\makecell[c]{IM}} & \colorbox{orange!71.634}{\strut unfortunately} & \colorbox{blue!0.046}{\strut ,} & \colorbox{blue!0.466}{\strut it} & \colorbox{blue!11.05}{\strut 's} & \colorbox{orange!59.83500000000001}{\strut not} & \colorbox{blue!96.37299999999999}{\strut \textcolor{white}{silly}} & \colorbox{orange!55.919}{\strut fun} & \colorbox{blue!2.645}{\strut unless} & \colorbox{blue!0.031}{\strut you} & \colorbox{blue!16.762}{\strut enjoy} & \colorbox{orange!20.22}{\strut really} & \colorbox{orange!100.0}{\strut bad} & \colorbox{orange!43.230000000000004}{\strut movies} & \colorbox{blue!0.27299999999999996}{\strut .} \\
 & 0.97455 & -0.00063 & -0.00634 & -0.15033 & 0.81403 & -1.31111 & 0.76075 & -0.03599 & -0.00042 & -0.22804 & 0.27508 & 1.36045 & 0.58812 & -0.00371 \\ \hline

\multirow{2}{*}{\makecell[c]{IM \\ \textit{modified}}} & \colorbox{orange!69.002}{\strut unfortunately} & \colorbox{blue!0.029}{\strut ,} & \colorbox{blue!0.316}{\strut it} & \colorbox{blue!14.588999999999999}{\strut 's} & \colorbox{orange!14.515}{\strut not} & \colorbox{blue!69.00800000000001}{\strut \textcolor{white}{silly}} & \colorbox{blue!0.12}{\strut fun} & \colorbox{orange!15.539}{\strut unless} & \colorbox{orange!0.015}{\strut you} & \colorbox{blue!19.442999999999998}{\strut enjoy} & \colorbox{orange!14.621999999999998}{\strut really} & \colorbox{orange!100.0}{\strut bad} & \colorbox{orange!32.35}{\strut movies} & \colorbox{blue!0.217}{\strut .} \\
 & 1.6679 & -0.00071 & -0.00764 & -0.35265 & 0.35085 & -1.66804 & -0.0029 & 0.37561 & 0.00036 & -0.46997 & 0.35344 & 2.41716 & 0.78194 & -0.00525 \\ \hline

\end{tabular}%
}
\caption{
Color map:
\colorbox{blue!100}{\strut \textcolor{white}{negative -1}}, neutral 0, \colorbox{orange!100}{\strut positive +1}.
Attribution maps derived from both versions of IM have a high Pearson correlation $\rho = 0.802$.
}
\label{fig:qualitative_examples_sst2_1d}
\end{figure*}

\begin{figure*}[ht]
\centering
\def\arraystretch{1.5}
\resizebox{\textwidth}{!}{%
\begin{tabular}{|c|c|c|c|c|c|c|c|c|c|c|c|c|c|c|c|c|}
\hline
\multicolumn{17}{|l|}{\cellcolor{white} \textbf{SST-2} example.~ Groundtruth: \class{positive} \& ~Prediction: \class{negative} (Confidence: 0.7999)} \\
\hline
\multirow{2}{*}{\makecell[c]{IM}} & \colorbox{blue!80.95}{\strut \textcolor{white}{intriguing}} & \colorbox{blue!26.457000000000004}{\strut documentary} & \colorbox{blue!52.051}{\strut which} & \colorbox{blue!1.247}{\strut is} & \colorbox{orange!4.478}{\strut emotionally} & \colorbox{orange!90.82000000000001}{\strut diluted} & \colorbox{orange!19.056}{\strut by} & \colorbox{orange!0.699}{\strut focusing} & \colorbox{orange!0.013}{\strut on} & \colorbox{orange!0.681}{\strut the} & \colorbox{blue!7.127}{\strut story} & \colorbox{orange!19.362}{\strut 's} & \colorbox{orange!100.0}{\strut least} & \colorbox{orange!16.733}{\strut interesting} & \colorbox{blue!2.481}{\strut subject} & \colorbox{blue!1.6809999999999998}{\strut .} \\
 & -7.28604 & -2.3813 & -4.68492 & -0.11221 & 0.40301 & 8.17448 & 1.71521 & 0.06288 & 0.00117 & 0.06125 & -0.64145 & 1.74269 & 9.00071 & 1.50607 & -0.22335 & -0.15134 \\ \hline

\multirow{2}{*}{\makecell[c]{IM \\ \textit{modified}}} & \colorbox{blue!33.945}{\strut intriguing} & \colorbox{blue!9.602}{\strut documentary} & \colorbox{blue!20.416}{\strut which} & \colorbox{orange!2.393}{\strut is} & \colorbox{orange!34.888000000000005}{\strut emotionally} & \colorbox{orange!100.0}{\strut diluted} & \colorbox{orange!5.827}{\strut by} & \colorbox{orange!7.526}{\strut focusing} & \colorbox{blue!0.026}{\strut on} & \colorbox{orange!0.386}{\strut the} & \colorbox{blue!3.6999999999999997}{\strut story} & \colorbox{orange!22.528000000000002}{\strut 's} & \colorbox{orange!85.301}{\strut least} & \colorbox{orange!19.848}{\strut interesting} & \colorbox{blue!3.702}{\strut subject} & \colorbox{orange!0.272}{\strut .}  \\
 & -3.96954 & -1.1229 & -2.38742 & 0.27984 & 4.07982 & 11.69405 & 0.68146 & 0.88004 & -0.00308 & 0.04509 & -0.43266 & 2.63444 & 9.97514 & 2.32102 & -0.43297 & 0.03175 \\ \hline

\end{tabular}%
}
\caption{
Color map:
\colorbox{blue!100}{\strut \textcolor{white}{negative -1}}, neutral 0, \colorbox{orange!100}{\strut positive +1}.
Attribution maps derived from both versions of IM have a high Pearson correlation $\rho = 0.950$.
}
\label{fig:qualitative_examples_sst2_1e}
\end{figure*}

\section{Sanity check result}
\label{sec:appendix_sanity_check}

\begin{table}[ht] 
\centering\small
\setlength\tabcolsep{3.5pt}
\begin{tabular}{llcc}
\toprule
Criteria & Method & \multicolumn{1}{c}{SST-2} & \multicolumn{1}{c}{e-SNLI} \\
\toprule
\multirow{2}{*}{\makecell[l]{(a) \% tokens \\ changing sign}} & \looEmpty & \textbf{71.41} $\pm$ 17.12 & \textbf{56.07} $\pm$ 21.82 \\
\cmidrule(l{2pt}r{2pt}){2-4}
& \im & 62.27 $\pm$ 17.75 & 49.57 $\pm$ 20.35 \\
\cmidrule(l{2pt}r{2pt}){1-4}
\multirow{2}{*}{\makecell[l]{(b) Average\\absolute of\\differences\\}} & \looEmpty & \textbf{0.46} $\pm$ 0.18 & \textbf{0.26} $\pm$ 0.14 \\
\cmidrule(l{2pt}r{2pt}){2-4}
& \im & 0.31 $\pm$ 0.12 & 0.16 $\pm$ 0.12 \\
\bottomrule
\end{tabular}
\caption{
The percentage (\%) of token (a) whose attribution scores change signs and (b) the average of absolute differences in attribution magnitude after classifier randomization (higher is better).
IM is consistently more insensitive than \looEmpty in both SST-2 and e-SNLI.
}
\label{table:sanity}
\end{table}


\begin{table*}[ht] 
\centering\small
\setlength\tabcolsep{2pt}
\resizebox{1\textwidth}{!}{
\begin{NiceTabular}{lccc|ccc|ccc}
\toprule
\multicolumn{1}{l}{Accuracy $\downarrow$} & \multicolumn{3}{c}{ROAR} & \multicolumn{3}{c}{\roarBert} & \multicolumn{3}{c}{\roarBertSst} \\
\cmidrule(l{2pt}r{2pt}){2-10}
Method & 10\% & 20\% & 30\% & 10\% & 20\% & 30\% & 10\% & 20\% & 30\% \\
\toprule
(a) \lime & 75.51 $\pm$ 0.55 & 75.30 $\pm$ 0.80 & 77.45 $\pm$ 0.70 & 78.14 $\pm$ 0.54 & 73.44 $\pm$ 0.65 & 70.57 $\pm$ 0.56 & 78.83 $\pm$ 1.28 & 74.47 $\pm$ 0.67 & 72.18 $\pm$ 1.02 \\
\cmidrule(l{2pt}r{2pt}){1-10}
(b) \limeBert & \textbf{73.99} $\pm$ 0.74 & 72.22 $\pm$ 0.73 & 70.82 $\pm$ 0.86 & \textbf{74.13} $\pm$ 0.72 & 70.44 $\pm$ 0.86 & 70.48 $\pm$ 0.63 & \textbf{75.78} $\pm$ 0.22 & 71.33 $\pm$ 1.04 & \textbf{68.76} $\pm$ 0.79 \\
\cmidrule(l{2pt}r{2pt}){1-10} 
(c) \limeBertSst & 74.15 $\pm$ 1.26 & \textbf{70.85} $\pm$ 0.89 & \textbf{70.48} $\pm$ 0.98 & 76.19 $\pm$ 0.91 & \textbf{69.77} $\pm$ 0.46 & \textbf{67.61} $\pm$ 0.53 & 76.08 $\pm$ 0.46 & \textbf{70.92} $\pm$ 0.64 & 71.08 $\pm$ 0.34 \\
\bottomrule
\end{NiceTabular}
}
\caption{
Dev-set mean accuracy (\%) of 5 models trained on the new SST-2 examples where $N$\% of highest-attribution words per example are removed (\ie ROAR), replaced via BERT (\ie \roarBert) or BERT finetuned on SST-2 to fill in a \mask token (\ie \roarBertSst).
The original accuracy when no tokens are removed (\ie $N=0$\%) is 92.62 $\pm$ 0.30.
On average, under three metrics, \limeBert (b) and \limeBertSst (c) are better, \ie lower mean accuracy, than \lime (a).
}
\label{table:lime_vs_limebert}
\end{table*}


\begin{figure*}[ht]
\centering\small 
\setlength\tabcolsep{2pt} 
\begin{tabular}{|l|l|}
\hline
\multicolumn{2}{|l|}{\cellcolor{white} \textbf{SST} example.~ Groundtruth: \class{positive}} \\
\hline
S & \footnotesize may not have generated many sparks , but with his affection for Astoria and its people \colorbox{yellow!100}{\strut he} \colorbox{yellow!100}{\strut has} \colorbox{yellow!100}{\strut given} \colorbox{yellow!100}{\strut his} \colorbox{yellow!100}{\strut tale} \colorbox{yellow!100}{\strut a} \colorbox{yellow!100}{\strut warm} \colorbox{yellow!100}{\strut glow} . \\
\hline
\addlinespace[2mm]
\hline
S$_{1}$ & \footnotesize may not have generated many sparks , but with his affection for Astoria and its people \hlborderRed{he} has \hlborderGreen{given} his tale \hlborderBlue{a} warm glow . \\
\hline
& \textbf{0.9494 \quad \hlborderRed{he}} ~~~~~~~~~~~ \textbf{0.9105 \quad \hlborderGreen{given}} ~~~~~~~~ \textbf{0.9632 \quad \hlborderBlue{a}} \\
& 0.0103 \quad it ~~~~~~~~~~~~~~ 0.0285 \quad lent ~~~~~~~~~~~~ 0.0270 \quad its \\
& 0.0066 \quad , ~~~~~~~~~~~~~~~ 0.0143 \quad gave ~~~~~~~~~~~ 0.0033 \quad another \\
\hline
\end{tabular}
\caption{
\bertBase often correctly predicts the masked tokens (denoted in {\hlborderRed{red}}, {\hlborderGreen{green}}, {\hlborderBlue{blue}} rectangles) and assigns a high likelihood to the tokens that are labeled  \colorbox{yellow!100}{\strut important} by humans in the SST \class{positive} example. 
In each panel, we show the top-3 tokens suggested by BERT and their associated likelihoods.
}
\label{fig:qualitative_examples_sst_1b}
\end{figure*}

\begin{figure*}[ht]
\centering\small 
\setlength\tabcolsep{1.5pt} 
\begin{tabular}{|l|l|}
\hline
\multicolumn{2}{|l|}{\cellcolor{white} \textbf{SST} example.~ Groundtruth: \class{negative}} \\
\hline
S & \footnotesize Villeneuve spends \colorbox{yellow!100}{\strut too} \colorbox{yellow!100}{\strut much} \colorbox{yellow!100}{\strut time} \colorbox{yellow!100}{\strut wallowing} \colorbox{yellow!100}{\strut in} \colorbox{yellow!100}{\strut Bibi} \colorbox{yellow!100}{\strut 's} \colorbox{yellow!100}{\strut generic} \colorbox{yellow!100}{\strut angst} ( there are a lot of shots of her gazing out windows ) . \\
\hline
\addlinespace[2mm]
\hline
S$_{1}$ & \footnotesize Villeneuve spends too \hlborderRed{much} \hlborderGreen{time} wallowing \hlborderBlue{in} Bibi 's generic angst ( there are a lot of shots of her gazing out windows ) . \\
\hline
& \textbf{0.9987 \quad \hlborderRed{much}} ~~~~~~ \textbf{0.9976 \quad \hlborderGreen{time}} ~~~~~~~~ \textbf{0.9675 \quad \hlborderBlue{in}} \\
& 0.0011 \quad little ~~~~~~~~~ 0.0005 \quad money ~~~~~~ 0.0066 \quad with \\
& 0.0001 \quad some ~~~~~~~~ 0.0003 \quad space ~~~~~~~~ 0.0062 \quad on \\
\hline
\end{tabular}

\caption{
\bertBase often correctly predicts the masked tokens (denoted in {\hlborderRed{red}}, {\hlborderGreen{green}}, {\hlborderBlue{blue}} rectangles) and assigns a high likelihood to the tokens that are labeled  \colorbox{yellow!100}{\strut important} by humans in the SST \class{negative} example. 
In each panel, we show the top-3 tokens suggested by BERT and their associated likelihoods.
}
\label{fig:qualitative_examples_sst_1c}

\end{figure*}

\begin{figure*}[ht]
\centering\small 
\setlength\tabcolsep{2pt} 
\begin{tabular}{|l|l|}
\hline
\multicolumn{2}{|l|}{\cellcolor{white} \textbf{e-SNLI} example.~ Groundtruth: \class{entailment}} \\
\hline
P & \makecell[l]{\footnotesize  The \colorbox{yellow!100}{\strut two} \colorbox{yellow!100}{\strut farmers} are working on a piece of \colorbox{yellow!100}{\strut John} \colorbox{yellow!100}{\strut Deere} \colorbox{yellow!100}{\strut equipment} .} \\
\hline
H & \footnotesize \colorbox{yellow!100}{\strut John} \colorbox{yellow!100}{\strut Deere} equipment is being worked on by \colorbox{yellow!100}{\strut two} \colorbox{yellow!100}{\strut farmers} \\
\hline
\addlinespace[2mm]
\hline
P$_\text{1}$ & \makecell[l]{\footnotesize The two farmers are working on a piece of \hlborderRed{John} Deere \hlborderGreen{equipment}} \\
\hline
H$_\text{1}$ & \footnotesize \hlborderBlue{John} Deere equipment is being worked on by two farmers \\
\hline
& \textbf{0.9995 \quad \hlborderRed{john}} ~~~~ \textbf{0.9877 \quad \hlborderGreen{equipment}} ~~ \textbf{0.9711 \quad \hlborderBlue{john}} \\
& 0.0000 \quad johnny ~~ 0.0057 \quad machinery ~~~~ 0.0243 \quad the \\
& 0.0000 \quad henry ~~~~ 0.0024 \quad hardware ~~~~~~ 0.0005 \quad a \\
\hline
\end{tabular}
\caption{
\bertBase often correctly predicts the masked tokens (denoted in {\hlborderRed{red}}, {\hlborderGreen{green}}, {\hlborderBlue{blue}} rectangles) and assigns a high likelihood to the tokens that are labeled  \colorbox{yellow!100}{\strut important} by humans in the e-SNLI \class{entailment} example which contains a pair of premise (P) and hypothesis (H). 
In each panel, we show the top-3 tokens suggested by BERT and their associated likelihoods.
}
\label{fig:qualitative_examples_esnli_1a}
\end{figure*}

\begin{figure*}[ht]
\centering\small 
\setlength\tabcolsep{2pt} 
\begin{tabular}{|l|l|}
\hline
\multicolumn{2}{|l|}{\cellcolor{white} \textbf{e-SNLI} example.~ Groundtruth: \class{neutral}} \\
\hline
P & \footnotesize A man uses a projector to give a presentation .  \\
\hline
H & \footnotesize A man is giving a presentation in \colorbox{yellow!100}{\strut front} \colorbox{yellow!100}{\strut of} \colorbox{yellow!100}{\strut a} \colorbox{yellow!100}{\strut large} \colorbox{yellow!100}{\strut crowd} . \\
\hline
\addlinespace[2mm]
\hline
P$_\text{1}$ & \footnotesize A man uses a projector to give a presentation . \\
\hline
H$_\text{1}$ & \footnotesize A man is giving a presentation in \hlborderRed{front} \hlborderGreen{of} \hlborderBlue{a} large crowd . \\
\hline
& \textbf{1.0000 \quad \hlborderRed{front}} ~~~~~~~ \textbf{0.9999 \quad \hlborderGreen{of}} ~~~~~ \textbf{0.9993 \quad \hlborderBlue{a}} \\
& 0.0000 \quad view ~~~~~~~~~ 0.0000 \quad to ~~~~~~ 0.0005 \quad the \\
& 0.0000 \quad presence ~~ 0.0000 \quad with ~~ 0.0001 \quad another \\
\hline
\end{tabular}
\caption{
\bertBase often correctly predicts the masked tokens (denoted in {\hlborderRed{red}}, {\hlborderGreen{green}}, {\hlborderBlue{blue}} rectangles) and assigns a high likelihood to the tokens that are labeled  \colorbox{yellow!100}{\strut important} by humans in the e-SNLI \class{neutral} example which contains a pair of premise (P) and hypothesis (H). 
In each panel, we show the top-3 tokens suggested by BERT and their associated likelihoods.
}
\label{fig:qualitative_examples_esnli_1c}
\end{figure*}

\begin{figure*}[ht]
\centering\small 
\setlength\tabcolsep{2pt} 
\begin{tabular}{|l|p{0.92\linewidth}|}
\hline
\multicolumn{2}{|l|}{\cellcolor{white} \textbf{MultiRC} example.~ Groundtruth \& Prediction: \class{True} (confidence: 0.98)} \\
\hline
P & \footnotesize What causes a change in motion ? The application of a force . Any time an object changes motion , a force has been applied . In what ways can this happen ? Force can cause an object at rest to start moving . Forces can cause objects to speed up or slow down . Forces can cause a moving object to stop . Forces can also cause a change in direction . \colorbox{yellow!100}{\strut In short , forces cause changes in motion . The moving object may change its speed , its direction , or both .} We know that changes in motion require a force . We know that the size of the force determines the change in motion . How much an objects motion changes when a force is applied depends on two things . It depends on the strength of the force . It also depends on the objects mass . Think about some simple tasks you may regularly do . You may pick up a baseball . This requires only a very small force . \\
\hline
Q & \footnotesize What factors cause changes in motion of a moving object ? \\
\hline
A & \footnotesize The object 's speed , direction , or both speed and direction \\
\hline
\addlinespace[2mm]
\hline
P$_\text{1}$ & \footnotesize What causes a change in motion ? The application of a force . Any time an object changes motion , a force has been applied . In what ways can this happen ? Force can cause an object at rest to start moving . Forces can cause objects to speed up or slow down . Forces can cause a moving object to stop . Forces can also cause a change in direction . In short , forces cause changes in motion . The \hlborderRed{moving} object may \hlborderGreen{change} its speed , its direction , \hlborderBlue{or} both . We know that changes in motion require a force . We know that the size of the force determines the change in motion . How much an objects motion changes when a force is applied depends on two things . It depends on the strength of the force . It also depends on the objects mass . Think about some simple tasks you may regularly do . You may pick up a baseball . This requires only a very small force . \\
\hline
& \textbf{0.9927 \quad \hlborderRed{moving}} ~~~~~~~ \textbf{0.9891 \quad \hlborderGreen{change}} ~~~~~ \textbf{0.9995 \quad \hlborderBlue{or}} \\
& 0.0023 \quad moved ~~~~~~~~~~ 0.0033 \quad alter ~~~~~~~~~~~ 0.0004 \quad and \\
& 0.0016 \quad stationary ~~~~~ 0.0018 \quad affect ~~~~~~~~~ 0.0000 \quad etc \\
\hline
Q$_\text{1}$ & \footnotesize John Deere equipment is being worked on by two farmers \\
\hline
A$_\text{1}$ & \footnotesize The object 's speed , direction , or both speed and direction \\
\hline
\end{tabular}
\caption{
\bertBase often correctly predicts the masked tokens (denoted in {\hlborderRed{red}}, {\hlborderGreen{green}}, {\hlborderBlue{blue}} rectangles) and assigns a high likelihood to the tokens that are labeled  \colorbox{yellow!100}{\strut important} by humans in the MultiRC \class{True} example which contains a triplet of paragraph (P), question (Q) and answer (A). 
In each panel, we show the top-3 tokens suggested by BERT and their associated likelihoods.
}
\label{fig:qualitative_examples_multirc_1a}
\end{figure*}

\begin{figure*}[ht]
\centering\small 
\setlength\tabcolsep{2pt} 
\begin{tabular}{|l|p{0.92\linewidth}|}
\hline
\multicolumn{2}{|l|}{\cellcolor{white} \textbf{MultiRC} example.~ Groundtruth \& Prediction: \class{False} (confidence: 0.74)} \\
\hline
P & \footnotesize There have been many organisms that have lived in Earths past . Only a tiny number of them became fossils . Still , scientists learn a lot from fossils . \colorbox{yellow!100}{\strut Fossils are our best clues about the history of life on Earth .} \colorbox{yellow!100}{\strut Fossils provide evidence about life on Earth .} \colorbox{yellow!100}{\strut They tell us that life on Earth has changed over time .} Fossils in younger rocks look like animals and plants that are living today . Fossils in older rocks are less like living organisms . Fossils can tell us about where the organism lived . Was it land or marine ? Fossils can even tell us if the water was shallow or deep . Fossils can even provide clues to ancient climates . \\
\hline
Q & \footnotesize What are three things scientists learn from fossils ? \\
\hline
A & \footnotesize Who lived in prehistoric times \\
\hline
\addlinespace[2mm]
\hline
P$_\text{1}$ & \footnotesize There have been many organisms that have lived in Earths past . Only a tiny number of them became fossils . Still , scientists learn a lot from fossils . Fossils are our best clues about the history of \hlborderRed{life} on Earth . Fossils provide evidence about life on Earth . They tell us that life on \hlborderGreen{Earth} has changed over \hlborderBlue{\strut time} . Fossils in younger rocks look like animals and plants that are living today . Fossils in older rocks are less like living organisms . Fossils can tell us about where the organism lived . Was it land or marine ? Fossils can even tell us if the water was shallow or deep . Fossils can even provide clues to ancient climates . \\
\hline
& \textbf{0.9984 \quad \hlborderRed{life}} ~~~~~~~~ \textbf{0.9982 \quad \hlborderGreen{earth}} ~~~~~ \textbf{0.9980 \quad \hlborderBlue{\strut time}} \\
& 0.0004 \quad living ~~~~~ 0.0007 \quad mars ~~~~~~~ 0.0007 \quad millennia \\
& 0.0002 \quad things ~~~~~ 0.0002 \quad land ~~~~~~~ 0.0003 \quad history \\
\hline
Q$_\text{1}$ & \footnotesize What are three things scientists learn from fossils ? \\
\hline
A$_\text{1}$ & \footnotesize Who lived in prehistoric times \\
\hline
\end{tabular}
\caption{
\bertBase often correctly predicts the masked tokens (denoted in 
{\hlborderRed{red}},
{\hlborderGreen{green}},
{\hlborderBlue{blue}} rectangles)
and assigns a high likelihood to the tokens that are labeled  \colorbox{yellow!100}{\strut important} by humans in the MultiRC \class{False} example which contains a triplet of paragraph (P), question (Q) and answer (A). 
In each panel, we show the top-3 tokens suggested by BERT and their associated likelihoods.
}
\label{fig:qualitative_examples_multirc_1b}
\end{figure*}

\begin{figure*}[ht]
\centering\small 
\setlength\tabcolsep{2pt} 

\begin{tabular}{|l|l|}
\hline
\multicolumn{2}{|l|}{\cellcolor{white} \textbf{SST} example.~ Groundtruth \& Prediction: \class{negative} (confidence: 1.00)} \\
\hline

S & \footnotesize For starters , the story \colorbox{yellow!100}{\strut is} \colorbox{yellow!100}{\strut just} \colorbox{yellow!100}{\strut too} \colorbox{yellow!100}{\strut slim} . \\
\hline
\addlinespace[2mm]
\hline
S$_\text{IM}$ & \footnotesize For \colorbox{ProcessBlue!100}{\strut starters} , the \colorbox{ProcessBlue!100}{\strut story} is \colorbox{ProcessBlue!100}{\strut just} \colorbox{ProcessBlue!100}{\strut too} slim . \\
\hline
& IoU: 0.33, precision: 0.50, recall: 0.50 \\
\hline
S$_\text{LOO}$ & \footnotesize For starters , the story is \colorbox{WildStrawberry!100}{\strut just} \colorbox{WildStrawberry!100}{\strut too} \colorbox{WildStrawberry!100}{\strut slim} . \\
\hline
& IoU: \textbf{0.75}, precision: \textbf{1.00}, recall: \textbf{0.75} \\
\hline
\end{tabular}

\caption{
The set of \colorbox{WildStrawberry!100}{\strut explanatory words} given by \looEmpty covers 75\% of \colorbox{yellow!100}{\strut human} highlights with higher precision and IoU in the SST \class{negative} example while there are a half of \colorbox{ProcessBlue!100}{\strut tokens highlighted by \im} are in correlation with human explanations.
}
\label{fig:qualitative_examples_sst_2c}

\end{figure*}

\begin{figure*}[ht]
\centering\small 
\setlength\tabcolsep{2pt} 
\begin{tabular}{|l|l|}
\hline
\multicolumn{2}{|l|}{\cellcolor{white} \textbf{e-SNLI} example.~ Groundtruth \& Prediction: \class{contradiction} (confidence: 1.00)} \\
\hline
P & \footnotesize Two men are \colorbox{yellow!100}{\strut cooking} food together on the corner of the street . \\
\hline
H & \footnotesize The two men are \colorbox{yellow!100}{\strut running} in a race . \\
\hline
\addlinespace[2mm]
\hline
P$_\text{IM}$ & \footnotesize Two men are cooking food together on the corner of the street . \\
\hline
H$_\text{IM}$ & \footnotesize \colorbox{ProcessBlue!100}{\strut The} two men are \colorbox{ProcessBlue!100}{\strut running} in a \colorbox{ProcessBlue!100}{\strut race} . \\
\hline
& IoU: 0.25, precision: 0.33, recall: 0.50 \\
\hline
\addlinespace[2mm]
\hline
P$_\text{LOO}$ & \footnotesize Two men are \colorbox{WildStrawberry!100}{\strut cooking} food together on the corner of the street . \\
\hline
H$_\text{LOO}$ & \footnotesize The two \colorbox{WildStrawberry!100}{\strut men} are \colorbox{WildStrawberry!100}{\strut running} in a \colorbox{WildStrawberry!100}{\strut race} . \\
\hline
& IoU: \textbf{0.50}, precision: \textbf{0.50}, recall: \textbf{1.00} \\
\hline
\end{tabular}
\caption{
The set of \colorbox{WildStrawberry!100}{\strut explanatory words} given by \looEmpty covers \textbf{all} highlights (higher precision and IoU) that are important to \colorbox{yellow!100}{\strut human} in the e-SNLI \class{contradiction} example which contains a pair of premise (P) and hypothesis (H) while there are \textbf{a half} of \colorbox{ProcessBlue!100}{\strut tokens highlighted by \im} are in correlation with human explanations.
}
\label{fig:qualitative_examples_esnli_2b}
\end{figure*}

\begin{figure*}[ht]
\centering\small 
\setlength\tabcolsep{2pt} 
\begin{tabular}{|l|l|}
\hline
\multicolumn{2}{|l|}{\cellcolor{white} \textbf{e-SNLI} example.~ Groundtruth \& Prediction: \class{neutral} (confidence: 1.00)} \\
\hline
P & \footnotesize Woman in a dress standing in front of a line of a clothing line , with clothes hanging on the line . \\
\hline
H & \footnotesize Her dress is \colorbox{yellow!100}{\strut dark} \colorbox{yellow!100}{\strut blue} . \\
\hline
\addlinespace[2mm]
\hline
P$_\text{IM}$ & \footnotesize \colorbox{ProcessBlue!100}{\strut Woman} in \colorbox{ProcessBlue!100}{\strut a} \colorbox{ProcessBlue!100}{\strut dress} standing in front of a line of a clothing line , with clothes hanging on the line . \\
\hline
H$_\text{IM}$ & \footnotesize Her \colorbox{ProcessBlue!100}{\strut dress} is dark blue . \\
\hline
& IoU: 0.00, precision: 0.00, recall: 0.00 \\
\hline
\addlinespace[2mm]
\hline
P$_\text{LOO}$ & \footnotesize Woman in a \colorbox{WildStrawberry!100}{\strut dress} standing in front of a line of a clothing line , with clothes hanging on the line . \\
\hline
H$_\text{LOO}$ & \footnotesize \colorbox{WildStrawberry!100}{\strut Her} \colorbox{WildStrawberry!100}{\strut dress} \colorbox{WildStrawberry!100}{\strut is} \colorbox{WildStrawberry!100}{\strut dark} \colorbox{WildStrawberry!100}{\strut blue} . \\
\hline
& IoU: \textbf{0.33}, precision: \textbf{0.33}, recall: \textbf{1.00} \\
\hline
\end{tabular}
\caption{
The set of \colorbox{WildStrawberry!100}{\strut explanatory words} given by \looEmpty covers \textbf{all} highlights (higher precision and IoU) that are important to \colorbox{yellow!100}{\strut human} in the e-SNLI \class{neutral} example which contains a pair of premise (P) and hypothesis (H) while there are \textbf{none} \colorbox{ProcessBlue!100}{\strut tokens highlighted by \im} are in correlation with human explanations.
}
\label{fig:qualitative_examples_esnli_2c}
\end{figure*}

\begin{figure*}[ht]
\centering\small 
\setlength\tabcolsep{2pt} 
\begin{tabular}{|l|p{0.92\linewidth}|}
\hline
\multicolumn{2}{|l|}{\cellcolor{white} \textbf{MultiRC} example.~ Groundtruth \& Prediction: \class{True} (confidence: 0.90)} \\
\hline
P & \footnotesize \colorbox{yellow!100}{\strut There} \colorbox{yellow!100}{\strut have} \colorbox{yellow!100}{\strut been} \colorbox{yellow!100}{\strut many} \colorbox{yellow!100}{\strut organisms} \colorbox{yellow!100}{\strut that} \colorbox{yellow!100}{\strut have} \colorbox{yellow!100}{\strut lived} \colorbox{yellow!100}{\strut in} \colorbox{yellow!100}{\strut Earths} \colorbox{yellow!100}{\strut past} \colorbox{yellow!100}{\strut .} \colorbox{yellow!100}{\strut Only} \colorbox{yellow!100}{\strut a} \colorbox{yellow!100}{\strut tiny} \colorbox{yellow!100}{\strut number} \colorbox{yellow!100}{\strut of} \colorbox{yellow!100}{\strut them} \colorbox{yellow!100}{\strut became} \colorbox{yellow!100}{\strut fossils} \colorbox{yellow!100}{\strut .} Still , scientists learn a lot from fossils . Fossils are our best clues about the history of life on Earth . Fossils provide evidence about life on Earth . They tell us that life on Earth has changed over time . Fossils in younger rocks look like animals and plants that are living today . Fossils in older rocks are less like living organisms . Fossils can tell us about where the organism lived . Was it land or marine ? Fossils can even tell us if the water was shallow or deep . Fossils can even provide clues to ancient climates . \\
\hline
Q & \footnotesize What happened to some organisms that lived in Earth 's past ? \\
\hline
A & \footnotesize They became fossils . Others did not become fossils \\
\hline
\addlinespace[2mm]
\hline
P$_\text{IM}$ & \footnotesize There have been many organisms that have lived in \colorbox{ProcessBlue!100}{\strut Earths} past \colorbox{ProcessBlue!100}{\strut .} Only a tiny number of them \colorbox{ProcessBlue!100}{\strut became} \colorbox{ProcessBlue!100}{\strut fossils} . Still , scientists learn a lot from fossils . Fossils are our best clues about the history of life on Earth . Fossils provide evidence about life on Earth . They tell us that life on Earth has changed over time . Fossils in younger rocks look like animals and plants that are living today . Fossils in older rocks are less like living organisms . Fossils can tell us about where the organism lived . Was it land or marine ? Fossils can even tell us if the water was shallow or deep . Fossils can even provide clues to ancient climates . \\
\hline
Q$_\text{IM}$ & \footnotesize What happened to \colorbox{ProcessBlue!100}{\strut some} organisms that lived in Earth \colorbox{ProcessBlue!100}{\strut 's} past ? \\
\hline
A$_\text{IM}$ & \footnotesize They became fossils . \colorbox{ProcessBlue!100}{\strut Others} \colorbox{ProcessBlue!100}{\strut did} not become fossils \\
\hline
& IoU: 0.16, precision: 0.50, recall: 0.19 \\
\hline
\addlinespace[2mm]
\hline
P$_\text{LOO}$ & \footnotesize \colorbox{WildStrawberry!100}{\strut There} \colorbox{WildStrawberry!100}{\strut have} \colorbox{WildStrawberry!100}{\strut been} \colorbox{WildStrawberry!100}{\strut many} \colorbox{WildStrawberry!100}{\strut organisms} \colorbox{WildStrawberry!100}{\strut that} \colorbox{WildStrawberry!100}{\strut have} \colorbox{WildStrawberry!100}{\strut lived} \colorbox{WildStrawberry!100}{\strut in} \colorbox{WildStrawberry!100}{\strut Earths} \colorbox{WildStrawberry!100}{\strut past} \colorbox{WildStrawberry!100}{\strut .} \colorbox{WildStrawberry!100}{\strut Only} \colorbox{WildStrawberry!100}{\strut a} tiny \colorbox{WildStrawberry!100}{\strut number} \colorbox{WildStrawberry!100}{\strut of} \colorbox{WildStrawberry!100}{\strut them} \colorbox{WildStrawberry!100}{\strut became} \colorbox{WildStrawberry!100}{\strut fossils} \colorbox{WildStrawberry!100}{\strut .} Still , \colorbox{WildStrawberry!100}{\strut scientists} \colorbox{WildStrawberry!100}{\strut learn} a \colorbox{WildStrawberry!100}{\strut lot} from fossils . Fossils are our \colorbox{WildStrawberry!100}{\strut best} clues about the history of life on Earth . Fossils \colorbox{WildStrawberry!100}{\strut provide} evidence about life on Earth . They tell us that life on Earth has changed over time . Fossils in younger rocks \colorbox{WildStrawberry!100}{\strut look} like animals and plants that are living today . Fossils in older rocks are less like living organisms . Fossils can tell us about where the organism lived . Was it land or marine ? Fossils can even tell us if the water was shallow or deep . Fossils can even provide clues to ancient climates \colorbox{WildStrawberry!100}{\strut .} \\
\hline
Q$_\text{LOO}$ & \footnotesize \colorbox{WildStrawberry!100}{\strut What} \colorbox{WildStrawberry!100}{\strut happened} to \colorbox{WildStrawberry!100}{\strut some} organisms that lived in Earth 's \colorbox{WildStrawberry!100}{\strut past} ? \\
\hline
A$_\text{LOO}$ & \footnotesize They became fossils \colorbox{WildStrawberry!100}{\strut .} \colorbox{WildStrawberry!100}{\strut Others} \colorbox{WildStrawberry!100}{\strut did} not \colorbox{WildStrawberry!100}{\strut become} fossils \\
\hline
& IoU: \textbf{0.56}, precision: \textbf{0.57}, recall: \textbf{0.95} \\
\hline
\end{tabular}
\caption{
The set of \colorbox{WildStrawberry!100}{\strut explanatory words} given by \looEmpty covers 95\% of \colorbox{yellow!100}{\strut human} highlights with higher precision and IoU in the MultiRC \class{True} example which contains a triplet of paragraph (P), question (Q) and answer (A) while there are only few tokens given by \colorbox{ProcessBlue!100}{\strut \im} are in correlation with human explanations.
}
\label{fig:loo_vs_im_multirc}
\end{figure*}

\begin{figure*}[ht]
\centering\small 
\setlength\tabcolsep{2pt} 
\begin{tabular}{|l|p{0.92\linewidth}|}
\hline
\multicolumn{2}{|l|}{\cellcolor{white} \textbf{MultiRC} example.~ Groundtruth \& Prediction: \class{False} (confidence: 0.99)} \\
\hline
P & \footnotesize There have been many organisms that have lived in Earths past . Only a tiny number of them became fossils . Still , scientists learn a lot from fossils . Fossils are our best clues about the history of life on Earth . Fossils provide evidence about life on Earth . They tell us that life on Earth has changed over time . \colorbox{yellow!100}{\strut Fossils} \colorbox{yellow!100}{\strut in} \colorbox{yellow!100}{\strut younger} \colorbox{yellow!100}{\strut rocks} \colorbox{yellow!100}{\strut look} \colorbox{yellow!100}{\strut like} \colorbox{yellow!100}{\strut animals} \colorbox{yellow!100}{\strut and} \colorbox{yellow!100}{\strut plants} \colorbox{yellow!100}{\strut that} \colorbox{yellow!100}{\strut are} \colorbox{yellow!100}{\strut living} \colorbox{yellow!100}{\strut today} \colorbox{yellow!100}{\strut .} \colorbox{yellow!100}{\strut Fossils} \colorbox{yellow!100}{\strut in} \colorbox{yellow!100}{\strut older} \colorbox{yellow!100}{\strut rocks} \colorbox{yellow!100}{\strut are} \colorbox{yellow!100}{\strut less} \colorbox{yellow!100}{\strut like} \colorbox{yellow!100}{\strut living} \colorbox{yellow!100}{\strut organisms} \colorbox{yellow!100}{\strut .} Fossils can tell us about where the organism lived . Was it land or marine ? Fossils can even tell us if the water was shallow or deep . Fossils can even provide clues to ancient climates . \\
\hline
Q & \footnotesize What is a major difference between younger fossils and older fossils ? \\
\hline
A & \footnotesize Older rocks are rougher and thicker than younger fossils \\
\hline
\addlinespace[2mm]
\hline
P$_\text{IM}$ & \footnotesize There have been many organisms that have lived in Earths past . Only a tiny number of them became fossils . Still , \colorbox{ProcessBlue!100}{\strut scientists} learn a lot from fossils . Fossils are our best \colorbox{ProcessBlue!100}{\strut clues} about the history of life on Earth . Fossils provide evidence about life on Earth . They tell us that life on Earth has changed over time . Fossils in younger rocks look like animals and plants that are living today . Fossils in older rocks \colorbox{ProcessBlue!100}{\strut are} less like \colorbox{ProcessBlue!100}{\strut living} organisms . Fossils can tell us about where the organism lived . Was it \colorbox{ProcessBlue!100}{\strut land} or \colorbox{ProcessBlue!100}{\strut marine} ? Fossils can even tell us if the water was shallow or deep . Fossils can even provide clues to ancient climates . \\
\hline
Q$_\text{IM}$ & \footnotesize What is a major difference between younger fossils and older fossils ? \\
\hline
A$_\text{IM}$ & \footnotesize Older \colorbox{ProcessBlue!100}{\strut rocks} are \colorbox{ProcessBlue!100}{\strut rougher} and \colorbox{ProcessBlue!100}{\strut thicker} \colorbox{ProcessBlue!100}{\strut than} \colorbox{ProcessBlue!100}{\strut younger} fossils \\
\hline
& IoU: 0.06, precision: 0.18, recall: 0.08 \\
\hline
\addlinespace[2mm]
\hline
P$_\text{LOO}$ & \footnotesize There have been many \colorbox{WildStrawberry!100}{\strut organisms} that have lived in \colorbox{WildStrawberry!100}{\strut Earths} past \colorbox{WildStrawberry!100}{\strut .} Only a tiny number of them became fossils . Still , scientists learn a lot from fossils . \colorbox{WildStrawberry!100}{\strut Fossils} are our \colorbox{WildStrawberry!100}{\strut best} \colorbox{WildStrawberry!100}{\strut clues} about the history of \colorbox{WildStrawberry!100}{\strut life} on \colorbox{WildStrawberry!100}{\strut Earth} \colorbox{WildStrawberry!100}{\strut .} Fossils provide evidence about life on Earth . They tell us that \colorbox{WildStrawberry!100}{\strut life} \colorbox{WildStrawberry!100}{\strut on} Earth has changed over time . \colorbox{WildStrawberry!100}{\strut Fossils} \colorbox{WildStrawberry!100}{\strut in} younger \colorbox{WildStrawberry!100}{\strut rocks} look like \colorbox{WildStrawberry!100}{\strut animals} \colorbox{WildStrawberry!100}{\strut and} \colorbox{WildStrawberry!100}{\strut plants} \colorbox{WildStrawberry!100}{\strut that} \colorbox{WildStrawberry!100}{\strut are} \colorbox{WildStrawberry!100}{\strut living} \colorbox{WildStrawberry!100}{\strut today} \colorbox{WildStrawberry!100}{\strut .} \colorbox{WildStrawberry!100}{\strut Fossils} in older rocks \colorbox{WildStrawberry!100}{\strut are} less like \colorbox{WildStrawberry!100}{\strut living} \colorbox{WildStrawberry!100}{\strut organisms} \colorbox{WildStrawberry!100}{\strut .} \colorbox{WildStrawberry!100}{\strut Fossils} \colorbox{WildStrawberry!100}{\strut can} \colorbox{WildStrawberry!100}{\strut tell} \colorbox{WildStrawberry!100}{\strut us} \colorbox{WildStrawberry!100}{\strut about} \colorbox{WildStrawberry!100}{\strut where} \colorbox{WildStrawberry!100}{\strut the} \colorbox{WildStrawberry!100}{\strut organism} \colorbox{WildStrawberry!100}{\strut lived} \colorbox{WildStrawberry!100}{\strut .} \colorbox{WildStrawberry!100}{\strut Was} \colorbox{WildStrawberry!100}{\strut it} \colorbox{WildStrawberry!100}{\strut land} \colorbox{WildStrawberry!100}{\strut or} \colorbox{WildStrawberry!100}{\strut marine} \colorbox{WildStrawberry!100}{\strut ?} \colorbox{WildStrawberry!100}{\strut Fossils} \colorbox{WildStrawberry!100}{\strut can} \colorbox{WildStrawberry!100}{\strut even} \colorbox{WildStrawberry!100}{\strut tell} us \colorbox{WildStrawberry!100}{\strut if} the water was shallow or deep . \colorbox{WildStrawberry!100}{\strut Fossils} \colorbox{WildStrawberry!100}{\strut can} even provide \colorbox{WildStrawberry!100}{\strut clues} to \colorbox{WildStrawberry!100}{\strut ancient} \colorbox{WildStrawberry!100}{\strut climates} \colorbox{WildStrawberry!100}{\strut .} \\
\hline
Q$_\text{LOO}$ & \footnotesize What \colorbox{WildStrawberry!100}{\strut is} a \colorbox{WildStrawberry!100}{\strut major} difference between younger \colorbox{WildStrawberry!100}{\strut fossils} and older \colorbox{WildStrawberry!100}{\strut fossils} \colorbox{WildStrawberry!100}{\strut ?} \\
\hline
A$_\text{LOO}$ & \footnotesize Older \colorbox{WildStrawberry!100}{\strut rocks} \colorbox{WildStrawberry!100}{\strut are} \colorbox{WildStrawberry!100}{\strut rougher} \colorbox{WildStrawberry!100}{\strut and} thicker \colorbox{WildStrawberry!100}{\strut than} \colorbox{WildStrawberry!100}{\strut younger} fossils \\
\hline
& IoU: \textbf{0.22}, precision: \textbf{0.25}, recall: \textbf{0.67} \\
\hline
\end{tabular}
\caption{
The set of \colorbox{WildStrawberry!100}{\strut explanatory words} given by \looEmpty covers two thirds of \colorbox{yellow!100}{\strut human} highlights with higher precision and IoU in the MultiRC \class{False} example which contains a triplet of paragraph (P), question (Q) and answer (A) while there are two tokens given by \im are in correlation with human explanations.
}
\label{fig:loo_vs_im_multirc2}
\end{figure*}

\begin{figure*}[ht]
\centering\small
\setlength\tabcolsep{3pt} 

\begin{tabular}{|l|l|}
\hline
\multicolumn{2}{|l|}{\cellcolor{white} \textbf{SST} example.~ Groundtruth \& Prediction: \class{positive}} \\
\hline

S & \footnotesize Enormously \colorbox{yellow!100}{\strut entertaining} for moviegoers of \colorbox{yellow!100}{\strut any} \colorbox{yellow!100}{\strut age} . \\
\hline
\hline 
S$_{1}$ & \footnotesize \textbf{\st{Enormously}} entertaining for moviegoers of any age . \\
\hline
S$_{2}$ & \footnotesize Enormously \textbf{\st{entertaining}} for moviegoers of any age . \\
\hline
S$_{3}$ & \footnotesize Enormously entertaining \textbf{\st{for}} moviegoers of any age . \\
\hline
S$_{4}$ & \footnotesize Enormously entertaining for \textbf{\st{moviegoers}} of any age . \\
\hline
S$_{5}$ & \footnotesize Enormously entertaining for moviegoers \textbf{\st{of}} any age . \\
\hline
S$_{6}$ & \footnotesize Enormously entertaining for moviegoers of \textbf{\st{any}} age . \\
\hline
S$_{7}$ & \footnotesize Enormously entertaining for moviegoers of any \textbf{\st{age}} . \\
\hline
\end{tabular}

\caption{
When a word is \textbf{\st{removed}}, the predicted labels of all resulting sentences (S$_1$ to S$_7$) are still \class{positive} with a confidence score of 1.0.
}
\label{fig:ood}
\end{figure*}

\begin{figure*}[ht]
\centering\small 
\setlength\tabcolsep{2pt} 

\begin{tabular}{|l|l|c|}
\hline
\multicolumn{2}{|l|}{\cellcolor{white} \textbf{e-SNLI} example.~ Groundtruth: \class{entailment}} & Prediction \\
\hline

P & \footnotesize Two women having \colorbox{yellow!100}{\strut drinks} and \colorbox{yellow!100}{\strut smoking} \colorbox{yellow!100}{\strut cigarettes} at the bar . & \multirow{2}{*}{\makecell{entailment \\ (0.99)}} \\
\cline{1-2}
H & \footnotesize Two women are at a \colorbox{yellow!100}{\strut bar} . & \\
\hline
\addlinespace[2mm]
\hline
P$_\text{1}$ & \footnotesize \textbf{\st{Two}} women having drinks and smoking cigarettes at the bar . & \multirow{2}{*}{\makecell{entailment \\ (0.98)}} \\
\cline{1-2}
H$_\text{1}$ & \footnotesize Two women are at a bar . & \\
\hline
\hline
P$_\text{2}$ & \footnotesize Two \textbf{\st{women}} having drinks and smoking cigarettes at the bar . & \multirow{2}{*}{\makecell{\textbf{neutral} \\ (0.93)}} \\
\cline{1-2}
H$_\text{2}$ & \footnotesize Two women are at a bar . & \\
\hline
\hline
P$_\text{3}$ & \footnotesize Two women \textbf{\st{having}} drinks and smoking cigarettes at the bar . & \multirow{2}{*}{\makecell{entailment \\ (0.99)}} \\
\cline{1-2}
H$_\text{3}$ & \footnotesize Two women are at a bar . & \\
\hline
\hline
P$_\text{4}$ & \footnotesize Two women having \textbf{\st{drinks}} and smoking cigarettes at the bar . & \multirow{2}{*}{\makecell{entailment \\ (0.99)}} \\
\cline{1-2}
H$_\text{5}$ & \footnotesize Two women are at a bar . & \\
\hline
\hline
P$_\text{5}$ & \footnotesize Two women having drinks \textbf{\st{and}} smoking cigarettes at the bar . & \multirow{2}{*}{\makecell{entailment \\ (0.99)}} \\
\cline{1-2}
H$_\text{5}$ & \footnotesize Two women are at a bar . & \\
\hline
\hline
P$_\text{6}$ & \footnotesize Two women having drinks and \textbf{\st{smoking}} cigarettes at the bar . & \multirow{2}{*}{\makecell{entailment \\ (0.99)}} \\
\cline{1-2}
H$_\text{6}$ & \footnotesize Two women are at a bar . & \\
\hline
\hline
P$_\text{7}$ & \footnotesize Two women having drinks and smoking \textbf{\st{cigarettes}} at the bar . & \multirow{2}{*}{\makecell{entailment \\ (0.99)}} \\
\cline{1-2}
H$_\text{7}$ & \footnotesize Two women are at a bar . & \\
\hline
\hline
P$_\text{8}$ & \footnotesize Two women having drinks and smoking cigarettes \textbf{\st{at}} the bar . & \multirow{2}{*}{\makecell{entailment \\ (0.98)}} \\
\cline{1-2}
H$_\text{8}$ & \footnotesize Two women are at a bar . & \\
\hline
\hline
P$_\text{9}$ & \footnotesize Two women having drinks and smoking cigarettes at \textbf{\st{the}} bar . & \multirow{2}{*}{\makecell{entailment \\ (0.98)}} \\
\cline{1-2}
H$_\text{9}$ & \footnotesize Two women are at a bar . & \\
\hline
\hline
P$_\text{10}$ & \footnotesize Two women having drinks and smoking cigarettes at the \textbf{\st{bar}} . & \multirow{2}{*}{\makecell{entailment \\ (0.97)}} \\
\cline{1-2}
H$_\text{10}$ & \footnotesize Two women are at a bar . & \\
\hline
\hline
P$_\text{11}$ & \footnotesize Two women having drinks and smoking cigarettes at the bar \textbf{\st{.}} & \multirow{2}{*}{\makecell{entailment \\ (0.99)}} \\
\cline{1-2}
H$_\text{11}$ & \footnotesize Two women are at a bar . & \\
\hline
\addlinespace[2mm]
\hline
P$_\text{12}$ & \footnotesize Two women having drinks and smoking cigarettes at the bar . & \multirow{2}{*}{\makecell{entailment \\ (0.99)}} \\
\cline{1-2}
H$_\text{12}$ & \footnotesize \textbf{\st{Two}} women are at a bar . & \\
\hline
\hline
P$_\text{13}$ & \footnotesize Two women having drinks and smoking cigarettes at the bar . & \multirow{2}{*}{\makecell{entailment \\ (0.98)}} \\
\cline{1-2}
H$_\text{13}$ & \footnotesize Two \textbf{\st{women}} are at a bar . & \\
\hline
\hline
P$_\text{14}$ & \footnotesize Two women having drinks and smoking cigarettes at the bar . & \multirow{2}{*}{\makecell{entailment \\ (0.99)}} \\
\cline{1-2}
H$_\text{14}$ & \footnotesize Two women \textbf{\st{are}} at a bar . & \\
\hline
\hline
P$_\text{15}$ & \footnotesize Two women having drinks and smoking cigarettes at the bar . & \multirow{2}{*}{\makecell{entailment \\ (\textbf{0.84})}} \\
\cline{1-2}
H$_\text{15}$ & \footnotesize Two women are \textbf{\st{at}} a bar . & \\
\hline
\hline
P$_\text{16}$ & \footnotesize Two women having drinks and smoking cigarettes at the bar . & \multirow{2}{*}{\makecell{entailment \\ (0.97)}} \\
\cline{1-2}
H$_\text{16}$ & \footnotesize Two women are at \textbf{\st{a}} bar . & \\
\hline
\hline
P$_\text{17}$ & \footnotesize Two women having drinks and smoking cigarettes at the bar . & \multirow{2}{*}{\makecell{entailment \\ (\textbf{0.54})}} \\
\cline{1-2}
H$_\text{17}$ & \footnotesize Two women are at a \textbf{\st{bar}} . & \\
\hline
\hline
P$_\text{18}$ & \footnotesize Two women having drinks and smoking cigarettes at the bar . & \multirow{2}{*}{\makecell{entailment \\ (0.95)}} \\
\cline{1-2}
H$_\text{18}$ & \footnotesize Two women are at a bar \textbf{\st{.}} & \\
\hline
\end{tabular}

\caption{
The removal of each token in both premise and hypothesis in e-SNLI example which contains a pair of premise (P) and hypothesis (H) \textbf{infrequently change the prediction}.
Specifically, only the example of (P$_{2}$, H$_{2}$) shifted its prediction to \class{neutral} while the remaining partially-removed examples do not change their original prediction with high confidence score in parentheses.
}
\label{fig:qualitative_examples_esnli_3a}

\end{figure*}


\end{document}